\documentclass[10pt,journal,compsoc]{IEEEtran}

\usepackage{amsmath,amsfonts}
\usepackage{array}
\usepackage[caption=false,font=normalsize,labelfont=sf,textfont=sf]{subfig}
\usepackage{textcomp}
\usepackage{stfloats}
\usepackage{url}
\usepackage{verbatim}
\usepackage{graphicx}
\usepackage[nice]{nicefrac}
\usepackage[normalem]{ulem}

\usepackage[linesnumbered,ruled,vlined]{algorithm2e}

\usepackage{multirow}
\usepackage[pagebackref,breaklinks,colorlinks,citecolor=blue,linkcolor=blue,]{hyperref}

\ifCLASSOPTIONcompsoc
  \usepackage[nocompress]{cite}
\else
  \usepackage{cite}
\fi
\ifCLASSINFOpdf
\else
\fi
\begin{document}

\title{Mixture of Gaussian-distributed Prototypes with Generative Modelling for Interpretable and Trustworthy Image Recognition}

\author{Chong Wang, \IEEEmembership{Member,~IEEE},
        Yuanhong Chen,
        Fengbei Liu,
        Yuyuan Liu,
        Davis James McCarthy, \\
        Helen Frazer, 
        and Gustavo Carneiro
\IEEEcompsocitemizethanks{
\IEEEcompsocthanksitem Chong Wang and Yuanhong Chen are with the Australian Institute for Machine Learning (AIML), University of Adelaide, SA 5000, Australia. Chong Wang is also with the Department of Radiology, Stanford University, Stanford, CA 94305-5105, USA.
E-mail: chong.wang@adelaide.edu.au; yuanhong.chen@adelaide.edu.au
\IEEEcompsocthanksitem Fengbei Liu is with the School of Electrical and Computer Engineering, Cornell University and Cornell Tech, New York, NY 10044, USA. 
E-mail: fl453@cornell.edu
\IEEEcompsocthanksitem Yuyuan Liu is with the Department of Engineering Science, University of Oxford, OX1 2JD Oxford, U.K. 
E-mail: yuyuan.liu@adelaide.edu.au
\IEEEcompsocthanksitem Davis James McCarthy is with the St. Vincent’s Institute of Medical Research, Melbourne, VIC 3065, Australia, and also with the Melbourne Integrative Genomics, The University of Melbourne, Melbourne, VIC 3010, Australia.
E-mail: dmccarthy@svi.edu.au
\IEEEcompsocthanksitem Helen Frazer is with the St Vincent’s Hospital Melbourne, Melbourne, VIC 3002, Australia.
E-mail: helen.frazer@svha.org.au
\IEEEcompsocthanksitem Gustavo Carneiro is with the Centre for Vision, Speech and Signal Processing (CVSSP), University of Surrey, GU2 7XH Guildford, U.K., and also with the Australian Institute for Machine Learning (AIML), The University of Adelaide, Adelaide, SA 5000, Australia.
E-mail: g.carneiro@surrey.ac.uk
}
\thanks{
This work was supported by the Australian Government under the Medical Research Future Fund for the Transforming Breast Cancer Screening with Artificial Intelligence (BRAIx) Project under Grant MRFAI000090 and the U.K. Engineering and Physical Sciences Research Council through Grant EP/Y018036/1.
Chong Wang and Yuanhong Chen contributed equally to this work. 
(Corresponding author: Gustavo Carneiro.)}
}

\markboth{IEEE Transactions on Pattern Analysis and Machine Intelligence,~Vol.~XX, No.~XX}%
{Shell \MakeLowercase{\textit{et al.}}: Bare Demo of IEEEtran.cls for Computer Society Journals}

\IEEEtitleabstractindextext{%
\vspace{-5pt}
\begin{abstract}


Prototypical-part methods, e.g., ProtoPNet, enhance interpretability in image recognition by linking predictions to training prototypes, 
thereby offering intuitive insights into their decision-making. 
Existing methods, which rely on a point-based learning of prototypes, typically face two critical issues: 
1) the learned prototypes have limited representation power and are not suitable to detect Out-of-Distribution (OoD) inputs, reducing their decision trustworthiness; and 
2) the necessary projection of the learned prototypes back into the space of training images causes a drastic degradation in the predictive performance. 
Furthermore, current prototype learning adopts an aggressive approach that considers only the most active object parts during training, while overlooking sub-salient object regions which still hold crucial classification information. 
In this paper, we present a new generative paradigm to learn prototype distributions, termed as \textbf{M}ixture of \textbf{G}aussian-distributed \textbf{Proto}types (MGProto). 
The distribution of prototypes from MGProto enables both interpretable image classification and trustworthy recognition of OoD inputs. 
The optimisation of MGProto naturally projects the learned prototype distributions back into the training image space, thereby addressing the performance degradation caused by prototype projection. 
Additionally, we develop a novel and effective prototype mining strategy that considers not only the most active but also sub-salient object parts. 
To promote model compactness, we further propose to prune MGProto by removing prototypes with low importance priors. 
Experiments on CUB-200-2011, Stanford Cars, Stanford Dogs, and Oxford-IIIT Pets datasets show that MGProto achieves state-of-the-art image recognition and OoD detection performances,
while providing encouraging interpretability results. 
Code is available at https://github.com/cwangrun/MGProto.


\end{abstract}

}

\maketitle

\IEEEdisplaynontitleabstractindextext
\IEEEpeerreviewmaketitle

\section{Introduction}
\label{sec:introduction}
\IEEEPARstart{D}{eep} learning models~\cite{krizhevsky2017imagenet,lecun2015deep} show remarkable performance in computer vision tasks, but their complex network architectures, high non-linearity, and massive parameter spaces make their decision processes opaque~\cite{rudin2019stop,tjoa2020survey}. 
As a result, deep learning models may not be trustworthy, particularly in safety-critical domains where the consequences of model errors can be severe~\cite{tjoa2020survey,frazer2024comparison,kim2017interpretable,wang2022bowelnet}. 
To alleviate this issue, explainable artificial intelligence (XAI) has gained increasing focus, striving to develop interpretable strategies to explain the internal workings of deep learning models in a manner comprehensible to humans~\cite{koh2017understanding,alvarez2018towards,chen2019looks,bohle2022b,wang2023interpretable}.
Among them, the prototypical-part network (ProtoPNet)~\cite{chen2019looks} is an appealing grey-box approach that classifies images based on similarities of the test sample to the prototypes corresponding to image patches previously-seen in the training set. 
Such strategy relies on train-test sample associations for decision making, which resembles how humans reason according to cognitive psychological studies~\cite{aamodt1994case,yang2021multiple} revealing that humans use past cases as models when learning to solve problems. 
The success of ProtoPNet has motivated the development of its many variants, e.g., TesNet~\cite{wang2021interpretable}, ProtoTree~\cite{nauta2021neural}, and PIP-Net~\cite{nauta2023pip}.

\begin{figure}[t!]
    \centering
    \includegraphics[width=1.00\linewidth]{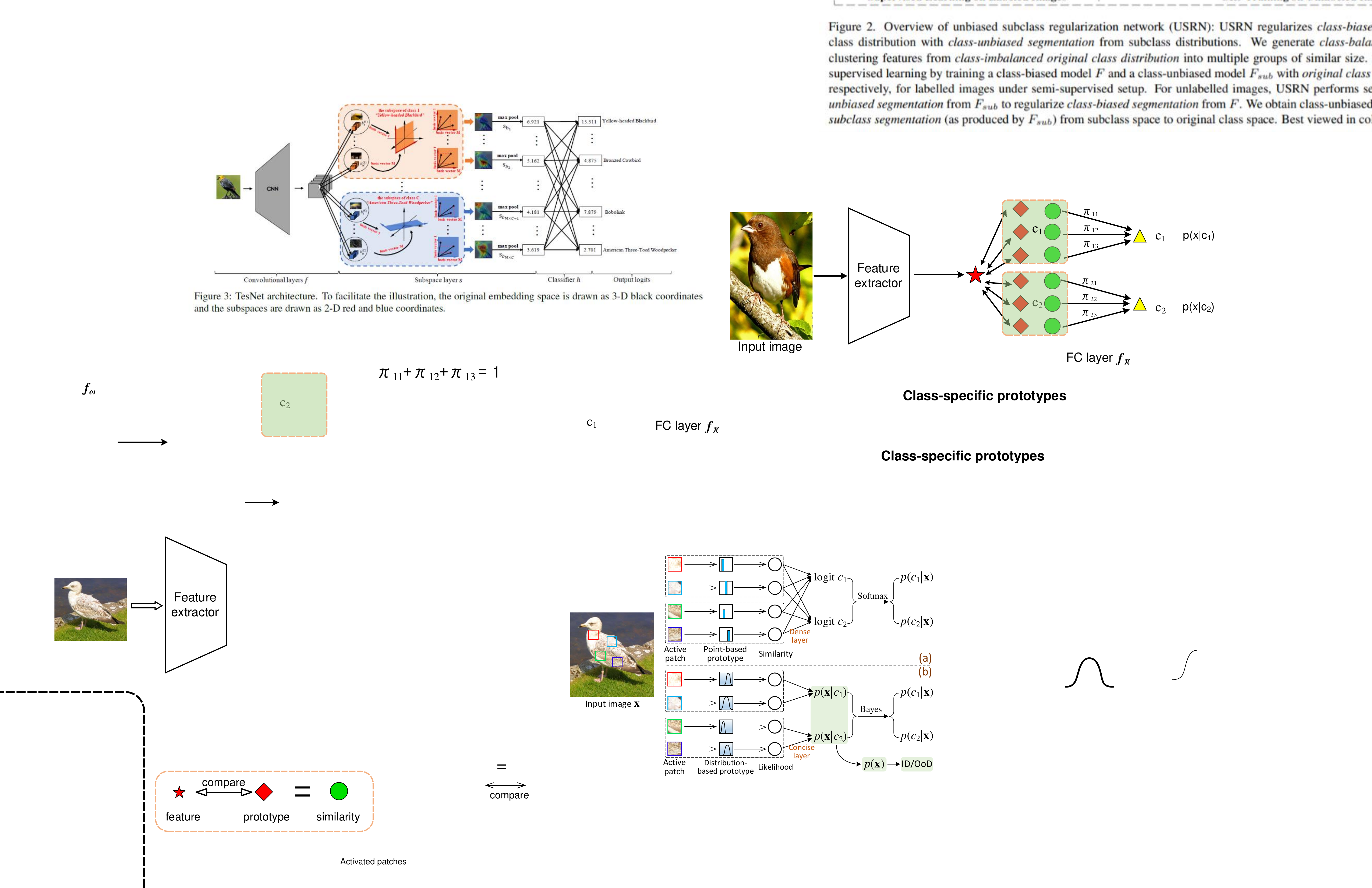}
    \vspace{-23pt}

    \caption{
    (a) Current prototypical-part networks are softmax-based discriminative classifiers, 
    forming a point-based learning of prototypes with limited representation power, which are challenged by the detection of OoD inputs. 
    (b) Our method learns a mixture of Gaussian-distributed prototypes with a generative modelling, 
    enabling not only interpretable image classification but also trustworthy recognition of OoD samples. 
    }
    \label{fig:introduction}
\end{figure}

\begin{figure}[t!]
    \centering
    \includegraphics[width=1.00\linewidth]{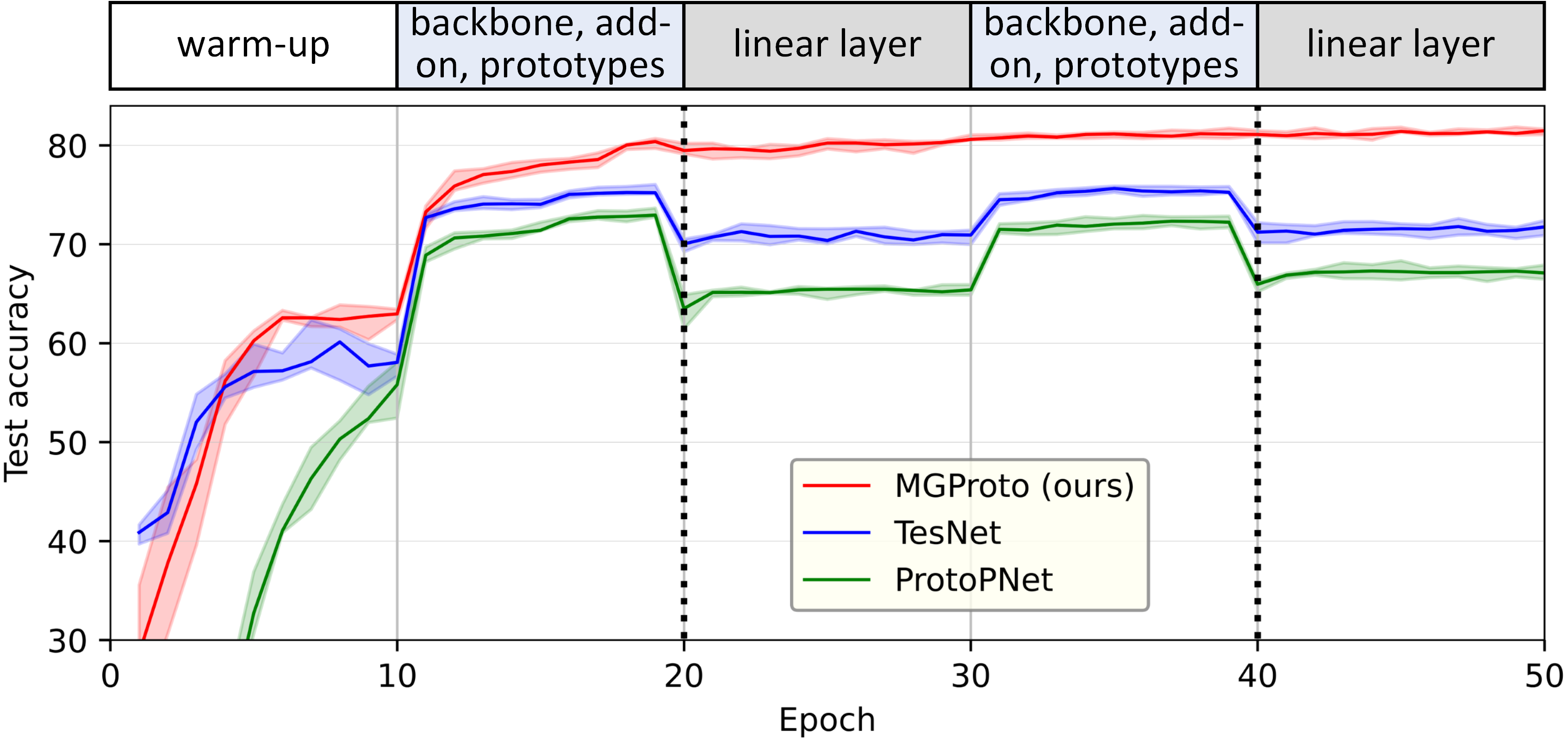}
    \vspace{-23pt}
    \caption{Current prototype-based  methods (e.g., ProtoPNet~\cite{chen2019looks} and TesNet~\cite{wang2021interpretable}) suffer from drastic performance degradation following the prototype replacement step (denoted by the dotted vertical lines) at each round of the multi-stage training, 
    whereas our MGProto method does not encounter this problem. 
    These curves are obtained from models trained on CUB-200-2011 using a ResNet34 backbone. 
    }
    \label{fig:replacement}
\end{figure}

Current prototypical part networks~\cite{chen2019looks,nauta2023pip,wang2023learning} depend on a discriminative classifier~\cite{bernardo2007generative} optimised with point-based learning techniques that train prototypes with specific values. 
Such a classifier produces logits (computed from a weighted sum of similarity scores), which are passed on to a softmax function to directly obtain the class probability given an input image, i.e., $p(c|\mathbf{x})$, as shown in Fig.~\ref{fig:introduction}(a). 
Though straightforward and easy to implement, the resulting prototypes from point-based learning 
have limited representation power and 
tends to suffer from drastic performance degradation (see Fig.~\ref{fig:replacement}) upon the prototype replacement\footnote{In the literature, the prototype replacement is also referred to as prototype “pushing” or “projection”. In this paper, we will use the terms replacement and projection interchangeably.} that is an indispensable step for grounding prototypes in the human-understandable space of training images~\cite{carmichael2024pixel}. 
Unfortunately, current methods either omit this replacement step~\cite{huang2023evaluation,xue2022protopformer,gautam2022protovae,kim2022vit} for better task accuracy, but at the cost of compromising interpretability given that their prototypes are no longer 
represented by actual training image patches, 
or include the replacement step without addressing the performance degradation issue~\cite{chen2019looks,wang2021interpretable,donnelly2022deformable,wang2023learning}.
Additionally, despite producing encouraging interpretable results, these discriminative classifiers miss an important explainability feature~\cite{mackowiak2021generative,serra2020input}, the recognition of
Out-of-Distribution (OoD) inputs. 
Ideally, an XAI system should be able to explain its predictions regarding In-Distribution (ID) samples (interpretable) and meanwhile identify anomalous OoD samples to abstain from classifying them to ensure the decision trustworthiness. 
Furthermore, the prototype learning in current methods is an aggressive\footnote{``Aggressive'' refers to the fact that the prototype's training process relies exclusively on the most salient object regions.}
approach that considers only the most salient object parts, 
disregarding important information available from less-salient object regions which could be helpful in the encoding of more difficult-to-learn visual features that have the potential to achieve improved classification.




Some issues mentioned above can be solved by generative models~\cite{bernardo2007generative} to learn prototype distributions with the class-conditional data density $p(\mathbf{x}|c)$, 
where the classification decision is made according to Bayes' theorem. 
Through the explicitly modelling of data densities, generative models are particularly suitable for detecting OoD samples, as evidenced by numerous studies~\cite{bernardo2007generative,ng2001discriminative,serra2020input}. 
By revisiting the existing prototype-based networks, 
we showcase that with proper modifications, the prototypes from current point-based learning techniques can be represented by generative Gaussian mixture models (GMM),
yielding our \textbf{M}ixture of \textbf{G}aussian-distributed \textbf{Proto}types (MGProto), as shown in Fig.~\ref{fig:introduction}(b), 
where each prototype is characterised by a rich GMM representation. 
A particularly beneficial implication of this new type of prototype representation is its effectiveness in detecting OoD samples. 
Interestingly, the learning of our GMM-based prototypes has a natural prototype projection step, which effectively addresses the performance degradation issue, as shown in Fig.~\ref{fig:replacement}. 
Furthermore, instead of considering only the most salient object parts in the prototype learning, 
we present an effective method to further mine prototypes from sub-salient object regions, by drawing inspiration from a strategic approach that counters the winning strategy depicted in the classic Tian Ji's horse-racing legend~\cite{shu2012generalized}. 
To allow model compactness, we  propose pruning our MGProto model by discarding Gaussian-distributed prototypes with low importance. 
To summarise, our major contributions are:
\begin{enumerate}
    \item We propose a new generative modelling of prototype distributions based on Gaussian mixture models, allowing both interpretable image classification and trustworthy recognition of OoD inputs. 
    \item We leverage MGProto's optimisation to seamlessly project the learned prototype distributions back into the training sample space, thereby mitigating the performance drop caused by prototype projection. 
    \item We present a novel and generic approach to enhance  prototype learning by mining prototypes from abundant less-salient object regions, inspired by the classic legend of Tian Ji’s horse-racing. 
    \item We introduce a method to compress our MGProto by adaptively pruning prototypical Gaussian components that hold low prototype importance.
\end{enumerate}
Experimental results on CUB-200-2011, Stanford Cars, Stanford Dogs, and Oxford-IIIT Pets show that our proposed MGProto outperforms current state-of-the-art (SOTA) methods in terms of both image recognition and OoD detection. Moreover, our MGProto also exhibits promising quantitative interpretability results.

\section{Related Work}
\label{sec:relatedwork}
\subsection{Interpretability with Prototypes}

Efforts to understand deep learning networks often face challenges if relying solely on post-hoc techniques like saliency maps~\cite{simonyan2013deep,zeiler2014visualizing,selvaraju2017grad,fang2019attention}, explanation surrogates~\cite{lundberg2017unified,zhang2018interpreting,shitole2021one}, and counterfactual examples~\cite{goyal2019counterfactual,teney2020learning,kenny2021generating}. 
These strategies fall short in explaining the network's reasoning process, yielding results that may be unreliable and risky~\cite{rudin2019stop}. 

On the other hand, prototype-based networks provide an appealing way to access the model's inner workings. 
These networks dissect a test image sample by finding prototypes and combining evidence from these prototypes to make a decision. 
ProtoPNet~\cite{chen2019looks} is the original work that adopts a number of class-specific prototypes for interpretable image recognition. 
Built upon it, TesNet~\cite{wang2021interpretable} introduces the orthonormal transparent basis concepts, and Deformable ProtoPNet~\cite{donnelly2022deformable} makes prototypes spatially-deformable to handle object variations. 
Other efforts extend prototypes to visual transformers~\cite{xue2022protopformer,kim2022vit}, K-nearest neighbour~\cite{ukai2023looks}, knowledge distillation~\cite{keswani2022proto2proto}, and ensemble interpretations~\cite{wang2023learning}.  To gauge the prototype-based interpretability, two objective metrics are proposed, i.e., consistency and stability scores~\cite{huang2023evaluation}. 
Also, some studies relax the class-specific constraint to allow a reduced number of prototypes~\cite{rymarczyk2021protopshare,rymarczyk2022interpretable,nauta2021neural,kim2022vit}. 
Recent works focus on refining semantically confounded or ambiguous prototypes~\cite{bontempelli2023concept,ma2024looks} in ProtoPNet. 
Promisingly, 
the prototype-based interpretable networks are being applied in critical tasks beyond the ones in computer vision~\cite{wang2022knowledge,wang2025cross,gerstenberger2023differentiable,wu2024cephalometric,wang2025progressive}.

The existing models above are trained with point-based learning techniques, yielding prototypes with limited representation power. 
As described in Section~\ref{sec:introduction}, the training for these prototypes involves a prototype projection into the training image space that causes drastic performance drop. 
Furthermore, it is challenging to recognise OoD inputs using these prototypes trained with point-based learning methods, thereby reducing the decision trustworthiness of the model. 
Even though PIP-Net~\cite{nauta2023pip}, trained with point-based learning techniques, attempts to detect OoD samples, its detection performance is still far from satisfactory. 



\subsection{Gaussian Mixture Model}

Gaussian Mixture Model (GMM)~\cite{reynolds2009gaussian} is a probabilistic approach to represent data distributions with a mixture of Gaussian components, which has been extensively explored in a variety of applications, such as semantic segmentation~\cite{wu2023sparsely,liang2022gmmseg}, point cloud registration~\cite{yuan2020deepgmr}, and image compression~\cite{cheng2020learned}.  
Recently, GMM has been used to explain inter-layer deep-learning features~\cite{xie2023joint} with an approach designed to produce post-hoc explanations
that fall short in explaining the network’s reasoning, which notably differs from our grey-box model with prototype-based interpretability. 
In this paper, we demonstrate that  previous point-based prototype learning methods in the ProtoPNet family can be formulated with GMMs to explicitly capture the underlying class-conditional data distributions.

For estimating the parameters of GMM, the Expectation–Maximisation (EM) algorithm~\cite{moon1996expectation} is typically utilised, involving iterative steps of evaluating the data responsibility using the current parameters (E-step) and maximising the data log-likelihood (M-step). 
An advantage of our approach is that the M-step of GMM training relies on  estimating the mean components of the Gaussian distributions with a weighted average of training samples, which naturally mitigates the performance degradation caused by the projection of point-based learned prototypes.
Also, while offering closed-form solutions, the standard EM algorithm often fails to ensure diverse Gaussian components, which results in potential prototype redundancy. 
We address this by incorporating a new constraint into the objective function of the M-step to encourage diversity of prototypes. 
\subsection{Tian Ji's Horse Racing Legend}


Tian Ji's horse racing~\cite{shu2012generalized} is a well-known Chinese legend taking place between two parties of horse-racing enthusiasts: the general Tian Ji and the King Wei of Qi. 
Each party has three available horses, categorised into three speed levels: fast, regular, and slow. 
At the same speed level, the King's horses are slightly faster than those of Tian. 
However, Tian’s horse in a superior level is able to beat King’s horse in an inferior level. 
They decide to hold a race with a total of three rounds, where in each round, both parties need to use a different horse. 
The party who wins the majority of the rounds wins the race. 
Initially, the King seems poised to win the race due to the faster speed of his horses at every level.
However, Sun Bin, one of the most renowned military strategists in ancient Chinese history, proposes Tian a winning strategy: in the first round, Tian's fast horse races against the King's regular horse; in the second round, Tian's regular horse races against the King's slow horse; and in the third round, 
Tian's slow horse races against the King's fast horse. 
In such a strategy, Tian will win the race by winning two rounds while losing only one, as shown in Fig.~\ref{fig:submining}(b). 

The mathematical principle of this victorious strategy achieved more than 2000 years ago is still studied in modern game theory~\cite{wang2012cooperative,leng2006game}. 
In this work, we take inspiration from this legend to build an infallible winning strategy for the King: 
if the regular and the slow horses of the King are trained to be faster than the Tian Ji's fast horse, 
then the King will always win the race by winning at each round. 
This infallible winning strategy of the King resembles our mining of prototypes from sub-salient object parts, elaborated in Section 3.4.
To the best of our knowledge, we are the first to employ the horse-racing strategy in XAI research.

\section{Our MGProto Method}

\begin{figure*}[t!]
    \centering
    \includegraphics[width=1.00\linewidth]{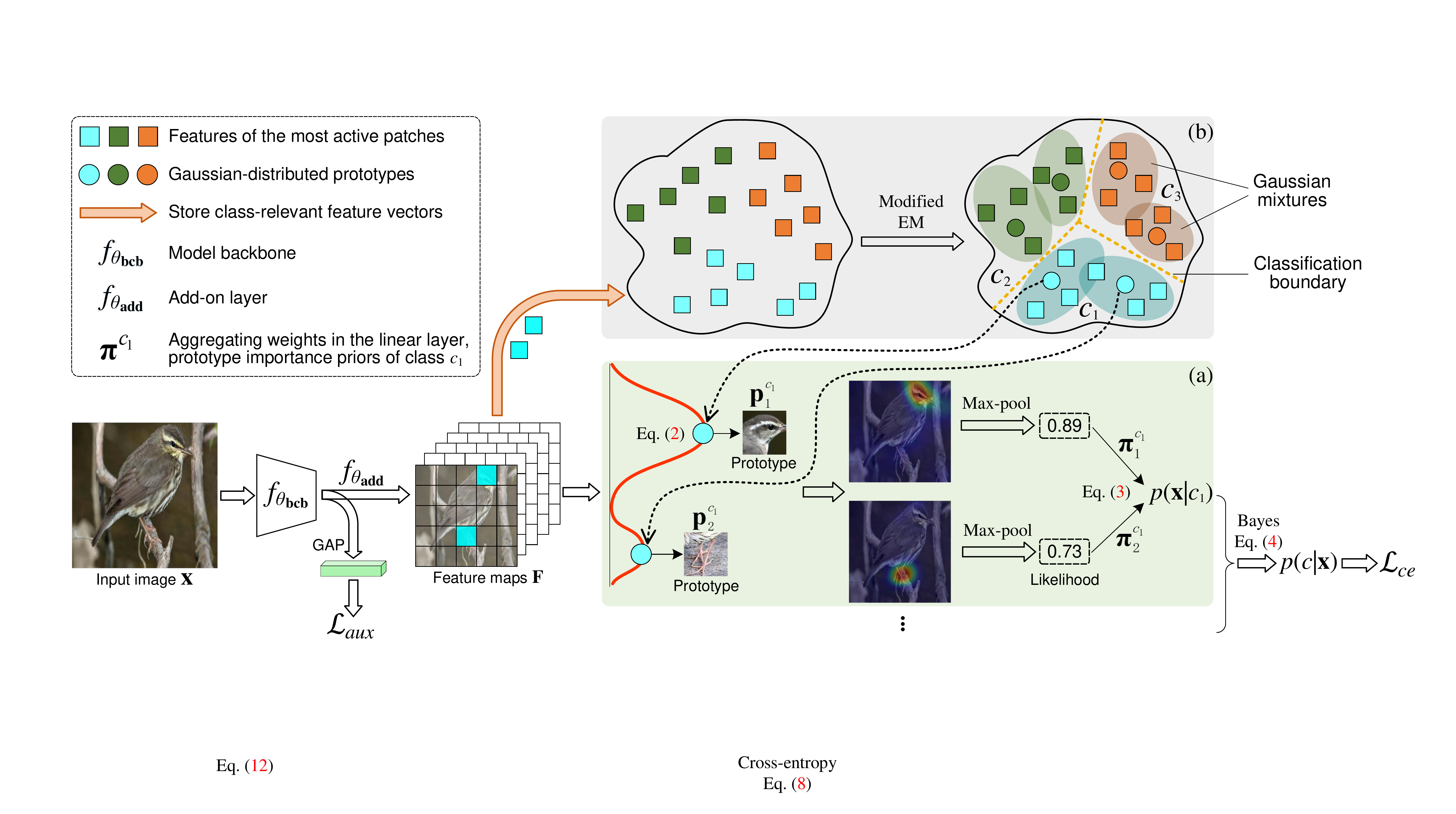}
    \vspace{-22pt}
    \caption{The overall framework of the MGProto method. For a given image $\mathbf{x}$, the model backbone (e.g., ResNet) extracts initial features $f_{\theta_{\textbf{bcb}}}(\mathbf{x})$ that are then fed to the add-on layer $f_{\theta_{\textbf{add}}}$ to obtain feature maps $\mathbf{F}$. 
    An auxiliary loss $\mathcal{L}_{aux}$ is applied on $f_{\theta_{\textbf{bcb}}}(\mathbf{x})$ to improve the backbone's feature extraction ability.
    (a) The case-based interpretation is achieved by fitting the feature representation $\mathbf{F}$ into the mixture of Gaussian-distributed prototypes, yielding the class-conditional data probability $p(\mathbf{x}|c)$ that enables the determination of whether the input is OoD. 
    Bayes' theorem is then used to derive the posterior class probability $p(c|\mathbf{x})$ for predicting the image category and computing the cross-entropy loss $\mathcal{L}_{ce}$. 
    For simplicity, here we show only 2 prototypes for class $c_1$, thus only 2 relevant features from the most active image patches are stored to the memory queue of class $c_1$.
    (b) For each class, the mixture of Gaussian-distributed prototypes is estimated by a modified EM algorithm to encourage prototype diversity. 
    }
    \label{fig:architecture}
\end{figure*}

This section starts with a brief description of the current ProtoPNet-based networks with point-based learning of prototypes in Section~\ref{sec:preliminaries}, serving as preliminaries for our proposed MGProto method illustrated in Fig.~\ref{fig:architecture}. 
Then, we detail our MGProto that represents prototypes with Gaussian mixture models (Section~\ref{sec:mixtureofgaussian}), which are learned via a modified EM algorithm to ensure prototype diversity (Section~\ref{sec:diverseprototypes}). 
Additionally, we present a novel and generic approach to mine prototypes from sub-salient object regions to improve classification, as elaborated in Section~\ref{sec:horseracing}.

\subsection{Preliminaries}
\label{sec:preliminaries}

To better introduce our distribution-based prototypes, we first briefly revisit the current point-based prototype learning methods, e.g., ProtoPNet~\cite{chen2019looks}. 
Let $\mathbf{x} \in \mathcal{X} \subset \mathbb{R}^{H \times W \times R}$ denote an image with size $H \times W$ and $R$ colour channels, and $\mathbf{y} \in \mathcal{Y} \subset \{0,1\}^C$ denote the one-hot image-level class label with $\mathbf{y}_c=1$ if the image class is $c$ and $\mathbf{y}_c=0$ otherwise. ProtoPNet  includes the following four steps: 

\noindent
1) \textbf{The embedding step} feeds an image $\mathbf{x}$ to a feature backbone $f_{\theta_{\textbf{bcb}}}:\mathcal{X} \to \mathcal{Z}$, parameterised by $\theta_{\textbf{bcb}}$, to extract initial features $\mathbf{z} \in \mathcal{Z} \subset \mathbb{R}^{\bar{H} \times \bar{W} \times \bar{D}}$, which is passed on to several add-on layers $f_{\theta_{\textbf{add}}}:\mathcal{Z} \to \mathcal{F}$, parameterised by $\theta_{\textbf{add}}$, to obtain feature maps $\mathbf{F} \in \mathcal{F} \subset \mathbb{R}^{\bar{H} \times \bar{W} \times D}$, where $\bar{H} \leq H$, $\bar{W} \leq W$ and $D$ denotes the number of feature channels. 

\noindent
2) \textbf{The prototype-activating step} uses a set of learnable prototypes $\mathcal{P}=\{\mathbf{p}_m\}_{m=1}^{M \times C}$ to represent prototypical object parts (e.g., tails and beaks from class ``bird'') in training images, 
where each of $C$ classes has $M$ prototypes and $\mathbf{p}_m \in \mathbb{R}^ {1 \times 1 \times D}$.
This step computes $M \times C$ similarity maps between the feature map $\mathbf{F}$ and prototypes $\mathcal{P}$, which are formulated as 
$\mathbf{S}_m^{(i,j)} = \mathsf{sim}(\mathbf{F}^{(i,j)}, \mathbf{p}_m)$, where 
$i \in \{1,...,\bar{H}\}$, 
$j \in \{1,...,\bar{W}\}$, $\mathbf{F}^{(i,j)} \in \mathbb{R}^ {1 \times 1 \times D}$, 
and $\mathsf{sim}(\cdot,\cdot)$ is a similarity metric (e.g., cosine similarity~\cite{donnelly2022deformable}). 
These similarity maps are then transformed into $M \times C$ similarity/activation scores from the max-pooling: $\mathbf{s}_m(\mathbf{x}) = \underset{i, j}\max \; \mathbf{S}_m^{(i,j)}$, where $m\in\{1,...,M \times C\}$.

\noindent
3) \textbf{The aggregating step} computes the logit of class $c$ 
by accumulating the similarity scores $\mathbf{s}_m(\mathbf{x})$ 
via a dense linear layer as in Fig.~\ref{fig:introduction}(a): $\mathbf{logit}^c = \sum_{m = 1}^{M \times C} \boldsymbol{\pi}_m^c \mathbf{s}_m(\mathbf{x})$, where $\boldsymbol{\pi}^c \in \mathbb{R}^{MC \times 1}$ is the layer's connection weight with class $c \in \{1,...,C\}$. 
A softmax function is applied to the output logits of all classes to predict the posterior class probability $p(c|\mathbf{x}; \theta_{\textbf{bcb}}, \theta_{\textbf{add}}, \mathcal{P}, \boldsymbol{\pi})$, where $\theta_{\textbf{bcb}}, \theta_{\textbf{add}}, \mathcal{P}$ and $\boldsymbol{\pi}$ are parameters of ProtoPNet. Current ProtoPNet-based networks are optimised with point-based learning techniques to minimise the cross-entropy error between the posterior class probability and ground-truth image label.  

\begin{figure}
\begin{minipage}[t]{0.497\linewidth}
    \centering
    \captionsetup{width=.95\linewidth}
    \includegraphics[width=1.0\textwidth]{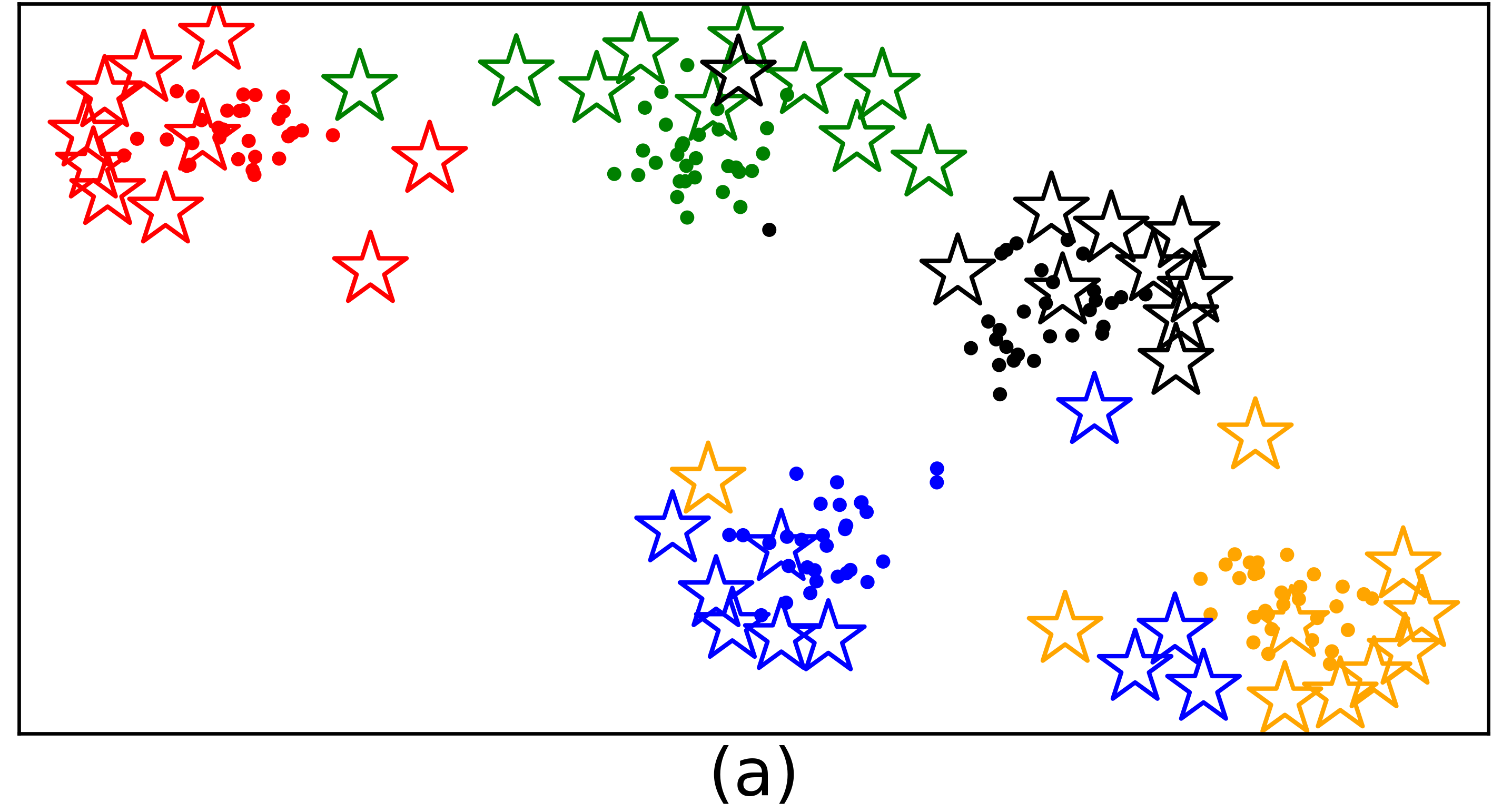}
    \label{fig:TSNE-ProtoPNet}
\end{minipage}
\begin{minipage}[t]{0.497\linewidth}
    \centering
    \captionsetup{width=.95\linewidth}
    \includegraphics[width=1.0\textwidth]{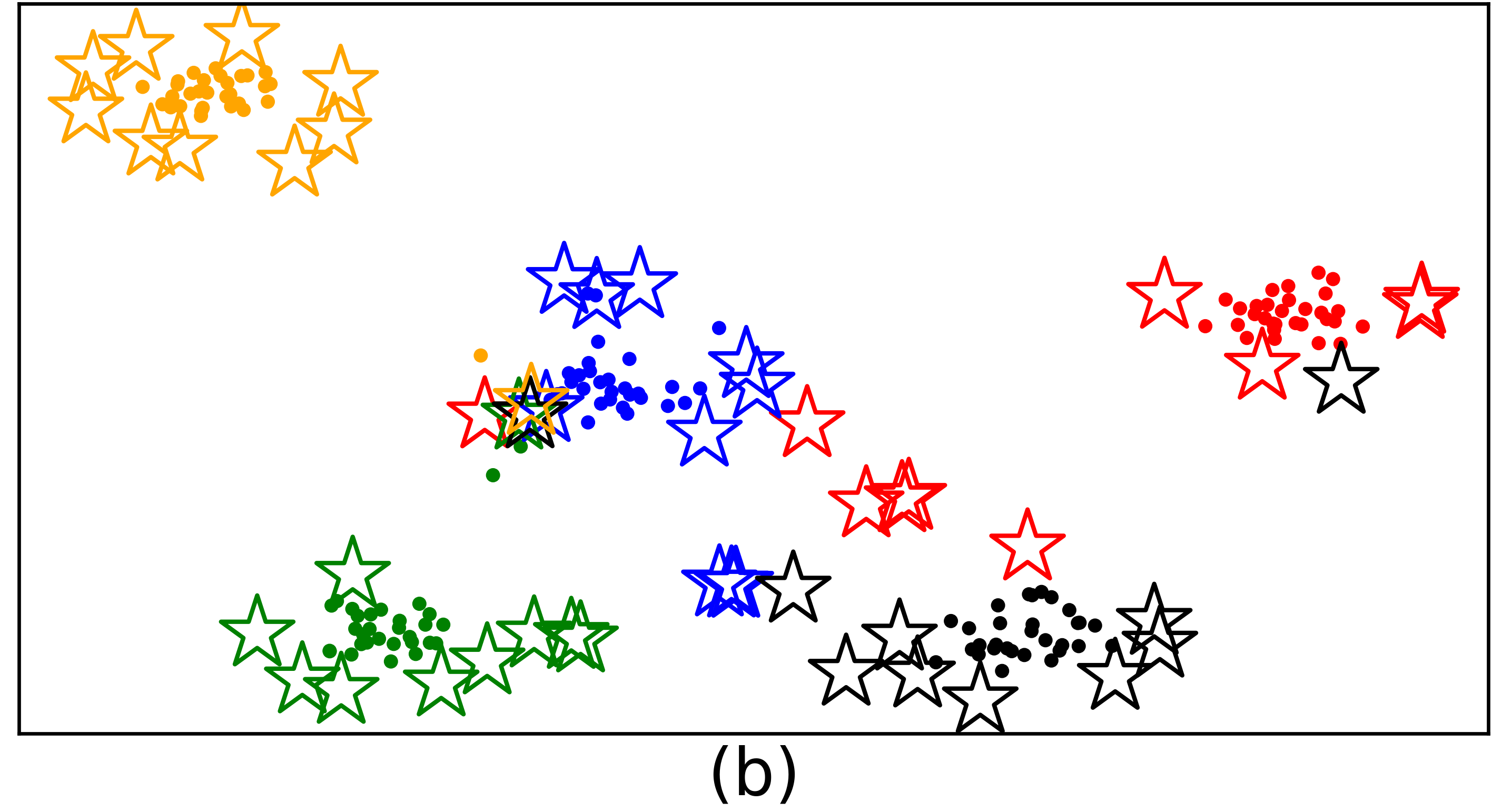}
    \label{fig:TSNE-TesNet}
\end{minipage}
\vspace{-23pt}
\caption{T-SNE representations of prototypes (stars) and the nearest training patch features (dots), from ProtoPNet (a) and TesNet (b), trained on CUB-200-2011. 
We show 5 random classes (out of 200) for better visualisation, where each colour denotes a different class. 
}
\label{fig:tsne}
\end{figure}

\noindent
4) \textbf{The replacement step}
grounds prototypes in the image space, so that prototypes are exactly represented by the actual training image patches~\cite{carmichael2024pixel}. 
This is reached by replacing each prototype with the latent feature of its nearest image patch of the same class in the training set~\cite{chen2019looks,donnelly2022deformable,wang2023learning}. Mathematically, for a prototype $\mathbf{p}_m$ of class $c$, we have: 
\begin{equation}
\ \mathbf{p}_m \leftarrow \arg\max_{\mathbf{f} \in \mathbf{F}_{a\in\{1,...,|\mathcal{D}_c|\}}} \mathsf{sim}(\mathbf{f}, \mathbf{p}_m),
\label{eq:prototype_replacement}
\end{equation}
where $\mathcal{D}_c$ is the set of training images from class $c$ and $\mathbf{f}$ is a latent feature vector of images in $\mathcal{D}_c$. 
Because the prototypes $\mathcal{P}$ are updated over training mini-batches, it is impractical to conduct the replacement step too often during training, given that it requires to search the most similar patch in the training set for each prototype. 
Usually, this step occurs every several training epochs, as shown in Fig.~\ref{fig:replacement}, and it causes drastic performance drops for ProtoPNet and TesNet. 
To help understand this issue, Fig.~\ref{fig:tsne} provides the T-SNE results of ProtoPNet and TesNet before the replacement, 
where we notice large distances or discrepancies between the point-based learned prototypes and their nearest training patch features. 
Such large distances explain why the sequential replacement causes drastic performance drops. 
A similar observation about this discrepancy can be found in~\cite{wang2023interpretable}.

ProtoPNet and its variants~\cite{wang2021interpretable,rymarczyk2022interpretable,donnelly2022deformable,wang2023learning} employ an iterative  multi-stage optimisation scheme (see Fig.~\ref{fig:replacement}) that alternates between: 
1) training model backbone, add-on layers, and prototypes; 
2) prototype replacement; 
and 3) training of the linear aggregating layer. 
After sufficient rounds, the performance degradation could be slightly alleviated, 
yet the final performance remains significantly deteriorated compared to the initial one.



\subsection{Mixture of Gaussian-distributed Prototypes}
\label{sec:mixtureofgaussian}

According to Section~\ref{sec:preliminaries}, current discriminative networks rely on point-based learned prototypes $\mathcal{P}$ that are used to compute similarity maps with $\mathbf{F}$.
Instead, we leverage a Gaussian distribution to model our distribution-based prototypes and obtain likelihood maps whose value at the spatial position $(i, j)$ denotes the confidence that the image patch fits the $m$-th prototype distribution of class $c$, as in: 
\begin{equation}
\resizebox{0.99\hsize}{!}{$
\begin{split}
    \mathbf{H}_m^{(i,j),c} & = \mathcal{N}(\mathbf{F}^{(i,j)}; \mathbf{p}^c_m, \mathbf{\Sigma}) \\
     & = \frac{1}{(2\pi)^{\frac{D}{2}} {|\mathbf{\Sigma}|}^{\frac{1}{2}}} e^{-\frac{1}{2} {(\mathbf{F}^{(i,j)}-\mathbf{p}^c_m)}^{\rm T} \mathbf{\Sigma}^{-1} {(\mathbf{F}^{(i,j)}-\mathbf{p}^c_m)} },
\end{split}
\label{eq:gaussian}
$}
\end{equation}
where the prototype $\mathbf{p}_m^c \in \mathbb{R}^{D}$ can be regarded as a mean of the Gaussian distribution, and $\mathbf{\Sigma} \in \mathbb{R}^{D \times D}$ is a constant diagonal covariance matrix with  diagonal value $\nicefrac{1}{2\pi}$, so that the likelihood has a range in $[0,1]$.
Note that we treat the covariance as constant, so our Gaussian-distributed prototypes do not introduce any extra parameters compared to existing point-based learning methods. 
We also apply max-pooling on the likelihood maps to obtain class-wise maximum likelihood scores: 
$\mathbf{h}^{c}_m(\mathbf{x}) = \underset{i, j}\max \; \mathbf{H}_m^{(i,j),c}$, for $m\in\{1,...,M\}$ and $c \in\{1,...,C\}$.

In alignment with previous prototype-based networks~\cite{chen2019looks,donnelly2022deformable,wang2023learning} that employ a set of $M$  prototypes to acquire a rich representation of a visual class, our MGProto also harnesses multiple prototype distributions for each class $c$, whose likelihood scores are accumulated by weighted sum in the aggregating layer, as shown in Fig.~\ref{fig:introduction}(b). 
This procedure naturally inspires us to derive the GMM formulation, which captures the class-conditional data density with a generative paradigm: 
\begin{equation}
\begin{split}
    p(\mathbf{x}|c) & = \sum_{m = 1}^{M} \boldsymbol{\pi}_m^c \mathbf{h}_m^c(\mathbf{x})  = \sum_{m = 1}^{M} \boldsymbol{\pi}_m^c \underset{i, j}\max \; \mathbf{H}_m^{(i,j), c} 
    \\
                   & = \sum_{m = 1}^{M} \boldsymbol{\pi}_m^c \underset{i, j}\max \; \mathcal{N}(\mathbf{F}^{(i,j)}; \mathbf{p}_m^c, \mathbf{\Sigma}),
\end{split}
\label{eq:class_conditional_prob}
\end{equation}
where $\boldsymbol{\pi}_m^c$ is the mixture weights in GMMs to quantify prototype importance, which are refered to as \textit{importance priors}. 
Also, $\boldsymbol{\pi}_m^c$ serves as the weights in the linear aggregating layer. 
From Eq.~(\ref{eq:class_conditional_prob}), the mixture model allows the prototype distributions of a class to collaboratively describe the underlying data density of that class. 
Meanwhile, these multiple prototype distributions in a mixture are enforced to compete against each other due to the inherent constraint in GMM: $\sum_{m = 1}^{M} \boldsymbol{\pi}_m^c = 1$ for each class $c$. 
Hence, we take advantage of this constraint to prune prototypes with low importance priors for promoting model compactness. 
It is worth noting that our MGProto utilises a concise aggregating layer (see Fig.~\ref{fig:introduction}(b)), represented by $\boldsymbol{\pi}^c \in \mathbb{R}^{M \times 1}$, which differs from current methods~\cite{chen2019looks,donnelly2022deformable,wang2023learning} using a dense layer denoted by $\boldsymbol{\pi}^c \in \mathbb{R}^{MC \times 1}$, as mentioned in Section~\ref{sec:preliminaries}.
The use of our concise layer not only introduces prototype-class connection sparsity to reduce the explanation size~\cite{nauta2023pip}, but also prevents the prediction of a class from being disturbed by prototypes of other classes~\cite{huang2023evaluation}.

Based on the well-known Bayes' theorem, the posterior class probability $p(c|\mathbf{x})$ is computed as 
\begin{equation}
    p(c|\mathbf{x}) = \frac{p(\mathbf{x}|c)p(c)}{\sum_{c' = 1}^{C} p(\mathbf{x}|c') p(c')}
    = \frac{p(\mathbf{x}|c)}{\sum_{c' = 1}^{C} p(\mathbf{x}|c')},
\label{eq:posterior}
\end{equation}
by assuming 
$p(c) = \frac{1}{C}$\footnote{The datasets used in our work have fairly balanced training samples across classes, making the prior $p(c)$ identical for all classes. 
For the imbalanced scenario, the prior can be estimated from the observed training data distribution. 
Alternatively, resampling strategies~\cite{estabrooks2004multiple} can be used during training, properly ensuring an identical prior.}, as commonly adopted in the literature~\cite{pearl1988probabilistic,rish2001empirical,peng2004augmenting}. 
Relying on this assumption, the core of MGProto is to accurately estimate the class-conditional data density $p(\mathbf{x}|c)$ modelled with the mixture of Gaussian-distributed prototypes through Eq.~(\ref{eq:class_conditional_prob}).

Equipped with class-conditioned distributions, 
our MGProto can identify OoD inputs and abstain from classifying them to ensure decision trustworthiness. 
This is achieved by marginalising  $p(\mathbf{x}|c)$ over all training classes to compute the overall data probability $p(\mathbf{x})$ that an input data belongs to the distribution of the training set: 
\begin{equation}
    p(\mathbf{x}) = \sum_{c = 1}^{C} p(\mathbf{x}|c)p(c) \propto \sum_{c = 1}^{C} p(\mathbf{x}|c).
\label{eq:ood}
\end{equation}
Based on Eq.~(\ref{eq:ood}), OoD inputs will have a low value for $p(\mathbf{x})$, which in practice means that $\mathbf{x}$ will be far from the prototypes of any of the classes and, consequently, will not fit well the prototype distributions of any class. 
Notice that Eq.~(\ref{eq:ood}) is an energy-like criteria, 
which has been adopted for OoD detection by discriminative classifiers~\cite{liu2020energy}, 
but not by generative models, as explored in our approach. 

The well-known expectation-maximisation (EM) algorithm~\cite{moon1996expectation} can be employed to estimate the prototype means $\mathbf{p}^c$ and importance priors $\boldsymbol{\pi}^c$ for each class $c$. 
Nevertheless, the standard EM algorithm does not guarantee diverse characteristics of the Gaussian-distributed prototypes, a crucial aspect for enhancing  the interpretability of prototypical-part  methods~\cite{wang2021interpretable,donnelly2022deformable,wang2023learning}. 
Therefore, in the next Section~\ref{sec:diverseprototypes}, we introduce a modified EM strategy to encourage diversity of the prototypes during their learning.


\subsection{Learning of Gaussian-distributed Prototypes}
\label{sec:diverseprototypes}


Section~\ref{sec:mixtureofgaussian} describes the prototype means $\mathbf{p}^c$ and importance priors $\boldsymbol{\pi}^c$ as parameters of the class-wise GMMs, defined in Eq.~(\ref{eq:class_conditional_prob}).
The accurate optimisation of GMM parameters typically 
depends on the whole training set~\cite{reynolds2009gaussian}.
However, modern deep-learning networks are often trained in a mini-batch fashion, 
where only a limited number of training data is provided.
That can harm GMM's predicative performance. 
To leverage a large set of feature representations for the GMM optimisation, we adopt the memory bank~\cite{chen2020simple} mechanism 
that has proven effective in maintaining contextual information~\cite{fan2022memory}.

The memory bank is defined as a class-wise queue $\mathcal{B}^c \in \mathbb{R}^{N \times D}$ that stores \textit{class-relevant} features from previously-processed training samples during learning, where $N$ is the memory capacity. 
As shown in Fig.~\ref{fig:architecture}, after a training image of class $c$ is processed by backbone and add-on layers, we obtain feature maps $\mathbf{F}$ consisting of $\bar{H} \times \bar{W}$ feature vectors. 
Since the training images usually contain background regions, not all these feature vectors are discriminative and relevant to the class $c$. 
We thus store only $M$ relevant feature vectors from the most active image patches into the queue $\mathcal{B}^c$ that correspond to the nearest feature vectors to the $M$ prototypes of class $c$. 
In other words, these $M$ feature vectors have the largest likelihood to  the $M$ prototype distributions. 
Note that the memory bank can be discarded after training, thus incurring no extra overhead for testing. 

Relying on the memory bank, our Gaussian-distributed prototypes $\{\mathbf{p}_m^c, \boldsymbol{\pi}_m^c \}_{m=1}^{M}$ can be estimated via a modified EM algorithm to accurately describe the underlying data densities. 
The E-step computes the responsibility that each feature $\mathbf{f}_n^c \in \mathcal{B}^c$, where $n \in \{1,...,N\}$, is generated by $m$-th Gaussian-distributed prototype of class $c$: 
\begin{equation}
    \gamma_{n, m}^c = \frac{\boldsymbol{\pi}_m^c \mathcal{N}(\mathbf{f}_n^c; \mathbf{p}_m^c, \mathbf{\Sigma})} {\sum_{m = 1}^{M} \boldsymbol{\pi}_m^c \mathcal{N}(\mathbf{f}_n^c; \mathbf{p}_m^c, \mathbf{\Sigma})}.
\label{eq:e-step-closeform}
\end{equation}

The standard EM algorithm provides a closed-form M-step solution~\cite{reynolds2009gaussian} to estimate the prototype means:
\begin{equation}
    \mathbf{p}_m^{c*} = \frac{1}{N_m} \sum_{n = 1}^{N} \gamma_{n, m}^c \mathbf{f}_n^c, \text{ where } N_m = \sum_{n = 1}^{N} \gamma_{n, m}^c.
\label{eq:m-step-closeform}
\end{equation}
However, the closed-form solution above does not guarantee the learning of diverse prototypes, which results in prototype redundancy and decreased performance in our experiments. 
Motivated by~\cite{kwok2012priors}, we modify the M-step by explicitly incorporating a prototype diversity constraint, with: 
\begin{equation}
\resizebox{.999\hsize}{!}{$
\begin{split}
    & \{\mathbf{p}_m^{c*} \}_{m=1}^{M} = \\ &\arg\max_{\{\mathbf{p}_m^c \}_{m=1}^{M}} \frac{1}{N} \sum_{n = 1}^{N} \sum_{m = 1}^{M} \gamma_{n, m}^c \log{\big(\boldsymbol{\pi}_m^c \mathcal{N}(\mathbf{f}_n^c; \mathbf{p}_m^c, \mathbf{\Sigma}) \big)} \\
    & \;\;\;\;\;\;\;\;\;\;\;\;\;\;\;\;\; - \frac{1}{{M(M-1)}} \sum_{m_1 = 1}^{M} \sum_{m_2 \neq m_1}^{M} e^{-||\mathbf{p}_{m_1}^c - \mathbf{p}_{m_2}^c||_2^2},
\end{split}
$}
\label{eq:m_step_loss}
\end{equation}
where the first term aims to maximise the log-likelihood over all data in the memory bank $\mathcal{B}^c$ 
and the second term improves prototype diversity by increasing within-class distances between prototype means. 
Note that Eq.~\eqref{eq:m_step_loss} no longer has closed-form solution, so we update it using gradient descent. 
The update of the prototype importance priors $\boldsymbol{\pi}_{m}^{c}$ can still rely on the closed-form solution:
\begin{equation}
    \boldsymbol{\pi}_{m}^{c*} = \frac{1}{N} \sum_{n = 1}^{N} \gamma_{n, m}^c.
\label{eq:m-step-pi}
\end{equation}

The GMM optimisation involves iterative loops between the E-step and M-step. 
Since our memory bank evolves progressively during training, only few loops (denoted by $L_{em}$) are sufficient for good EM convergence. 

According to Eq.~(\ref{eq:m-step-closeform}) and Eq.~(\ref{eq:m_step_loss}), 
our MGProto has a natural prototype projection step, happening during the M-step with the estimation of the prototype means as 
the weighted average of nearest training features stored in the memory queue, 
where the weight is the responsibility in Eq.~(\ref{eq:e-step-closeform}) that is determined by the relative affinity to the currently-estimated prototypes. 
Obviously, similar features to the prototypes will have larger weights and dominate the soft assignment process. 
By contrast, the prototype replacement in Eq.~(\ref{eq:prototype_replacement}) of existing methods are hard assignment of feature points, usually causing drastic performance drops. 


\subsection{Mining of Prototypes from Sub-salient Object Parts}
\label{sec:horseracing}

\begin{figure}[t!]
    \centering
    \includegraphics[width=1.00\linewidth]{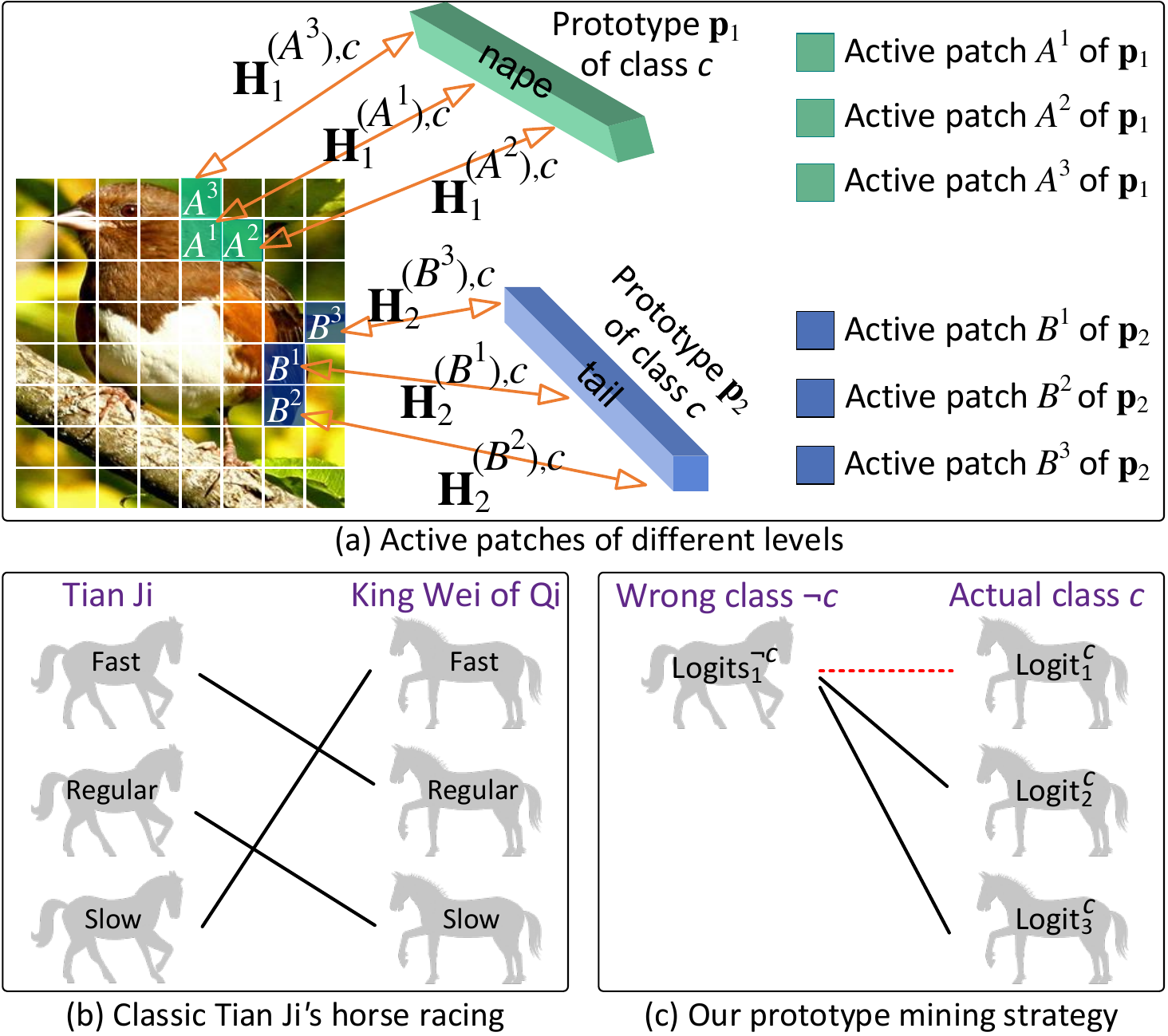}
    \vspace{-23pt}
    \caption{
    (a) Illustration of different levels of active patches for prototype mining. For clarity, here we suppose only two prototypes in each class and consider $T=3$ levels of active patches. 
    (b) Diagram of the classic Tian Ji's horse racing legend. 
    (c) Our proposed prototype mining strategy establishes $T-1$ mining competitions (solid lines). The standard classification supervision is represented by the dash line. 
    }
    \label{fig:submining}
\end{figure}

Similar to many existing methods that utilise a max-pooling operation in Eq.~(\ref{eq:class_conditional_prob}) to select the positions with the highest likelihood score, 
so far our prototype learning process aggressively harnesses only the most active patches (i.e., object parts) in an image. 
This means that during training, the learning process never involves the remaining less-active patches.
As a result, the prototypes will focus on only the most salient or discriminative object parts and overlook crucial information from other sub-salient object regions. 
Intuitively, these sub-salient regions encapsulate rich difficult-to-learn visual patterns that can be exploited to achieve improved classification.  
In Fig.~\ref{fig:submining}(a), we illustrate different levels of the active patches which have different likelihood scores for the same prototype.

In this paper, we present a novel and generic strategy, inspired by the ancient Tian Ji's horse racing legend~\cite{shu2012generalized}, to mine prototypes from sub-salient object regions.
As illustrated in Fig.~\ref{fig:submining}(b), 
the essence of this legend is that the ranking order of Tian’s horses should be wisely shifted, 
to allow Tian to win two out of the three rounds. 
We instead propose an infallible winning strategy for the King, consisting of training the King's regular and slow horses to be faster than Tian's fast horse.

We elaborate below how to formulate the new prototype mining strategy. 
Assume that we have a training image $\mathbf{x}$ labelled with class $c$. 
We consider $T$ sequential patch levels from the most to the least active, such as 
$\{A^1, A^2, ..., A^T\}$ of the prototype $\mathbf{p}_1$ and 
$\{B^1, B^2, ..., B^T\}$ of the prototype $\mathbf{p}_2$, 
as in Fig.~\ref{fig:submining}(a). 
According to Eq.~(\ref{eq:class_conditional_prob}), the predicted logit\footnote{In our generative MGProto, the class-conditional data probability $p(\mathbf{x}|c)$ serves as the model's prediction logits, but we still use the symbol $\mathbf{logit}$ here, given the generality of the proposed mining strategy to the existing discriminative-based prototypical-part networks.} for each patch level $t \in \{1, ...,T\}$ is computed as the weighted sum of likelihood scores $p(\mathbf{x}|c)$: $\mathbf{logit}^c_t = \boldsymbol{\pi}_{1}^{c} \times \mathbf{H}_{1}^{(A^{t}),c} + \boldsymbol{\pi}_{2}^{c} \times \mathbf{H}_{1}^{(B^{t}),c}$.
Obviously, we have $\mathbf{logit}^c_T < \mathbf{logit}^c_{T-1} <...< \mathbf{logit}^c_1$ because $\mathbf{H}_{1}^{(A^{T}),c} < \mathbf{H}_{1}^{(A^{T-1}),c} < ... < \mathbf{H}_{1}^{(A^{1}),c}$ and $\mathbf{H}_{1}^{(B^{T}),c} < \mathbf{H}_{1}^{(B^{T-1}),c} < ... < \mathbf{H}_{1}^{(B^{1}),c}$.
We showcase our prototype mining strategy in Fig.~\ref{fig:submining}(c), which is analogous to the legend. 
To propose the infallible winning strategy for the King, we regard: 
1) the most active logit of the wrong classes $\lnot c$, denoted by $\mathbf{logits}^{\lnot c}_1 \in \mathbb{R}^{C-1}$, as Tian Ji's fast horse; and 
2) the $T-1$ less active logits of the actual class $c$, denoted by $\mathbf{logit}^{c}_{t \in \{2,...,T\}}$, as the King's horses, except for the most active one denoted by $\mathbf{logit}^{c}_{1}$. 
Then, we devise $T-1$ mining competitions: $\mathbf{logits}^{\lnot c}_1$ vs. $\mathbf{logit}^{c}_{t \in \{2,...,T\}}$, 
which are supervised by the following mining loss with the goal of increasing $\mathbf{logit}^{c}_{t \in \{2,...,T\}}$:
\begin{equation}
    \mathcal{L}_{mn}(\mathbf{x}, \mathbf{y}) = \frac{1}{T-1} \sum_{t=2}^{T} \mathcal{L}_{ce}([\mathbf{logits}^{\lnot c}_{1}, \mathbf{logit}^{c}_{t}], \mathbf{y}), 
\label{eq:mining}
\end{equation}
where $\mathcal{L}_{ce}(\cdot)$ is the cross-entropy (CE) loss and $\mathbf{y}$ is the ground-truth label of $\mathbf{x}$. 
Here we employ $[\mathbf{logits}^{\lnot c}_{1}, \mathbf{logit}^{c}_{t}] \in \mathbb{R}^C$ to represent the combined mining logits corresponding to the solid lines in Fig.~\ref{fig:submining}(c), 
with $[\mathbf{logits}^{\lnot c}_{1}, \mathbf{logit}^{c}_{1}] \in \mathbb{R}^C$ denoting the standard classification logits, shown as the dash line, which is supervised by:  
\begin{equation}
    \mathcal{L}_{ce}(\mathbf{x}, \mathbf{y}) = \mathcal{L}_{ce}([\mathbf{logits}^{\lnot c}_{1}, \mathbf{logit}^{c}_{1}], \mathbf{y}), 
\label{eq:cross-entropy}
\end{equation}

Different from the legend in Fig.~\ref{fig:submining}(b), 
our prototype mining leverages all $\mathbf{logit}^{c}_{t \in \{2,...,T\}}$ with $T -1$ patch levels as in Eq.~(\ref{eq:mining}) to compete with the unique and most active $\mathbf{logits}^{\lnot c}_{1}$ to ensure an infallible winning strategy for the King, 
which also yields a stronger mining effect. 
To sum up, the major connections between our prototype mining and the classic horse-racing legend are: 
Tian Ji achieves victory through clever tactics in the legend. 
In contrast, our prototype mining approach devises a guaranteed winning strategy for the King, by training the King's regular and slower horses to ensure they surpass Tian Ji's fastest horse.
Our proposed prototype mining strategy can be readily applied to a wide range of prototypical-part networks, for improved classification. 



\subsection{Training Objective and Prototype Replacement}
\label{sec:traininglosses}

The overall training objective of our proposed MGProto method for a mini-batch $\mathcal{K}$ is defined as:
\begin{equation}
    \mathcal{L}_{total} = \frac{1}{|\mathcal{K}|} \sum_{(\mathbf{x},\mathbf{y}) \in \mathcal{K}} \mathcal{L}_{ce}(\mathbf{x},\mathbf{y}) + \lambda_1 \mathcal{L}_{mn}(\mathbf{x},\mathbf{y}) + \lambda_2 \mathcal{L}_{aux}(\mathbf{x}, \mathbf{y}),
\label{eq:totalloss}
\end{equation}
where 
$\lambda_1$ and $\lambda_2$ are hyper-parameters. 
In Eq.~\eqref{eq:totalloss}, the Proxy-Anchor~\cite{kim2020proxy} auxiliary loss $\mathcal{L}_{aux}(\cdot)$, proposed in the field of deep metric learning, 
is used to enhance the features extracted by the model backbone, as suggested by~\cite{ukai2023looks}. 
Specifically, we employ a Global Average Pooling (GAP) operator to condense the initial deep features $\mathbf{z} = f_{\theta_{\textbf{bcb}}}(\mathbf{x})$ into the embedding vector $\mathbf{e} = \mathsf{GAP}(\mathbf{z}) \in \mathbb{R}^{\bar{D}}$, where the loss is computed with:
\begin{equation}
\resizebox{.99\hsize}{!}{$
\begin{split}
    \mathcal{L}_{aux}(\mathbf{x}, \mathbf{y}) = \ & \frac{1}{|\mathcal{Q}^{+}|} \sum_{\mathbf{q} \in \mathcal{Q}^{+}} \log \Big(1 + \sum_{\mathbf{e} \in \mathcal{E}_{\mathbf{q}}^{+}} e^{-\alpha(\mathsf{sim}(\mathbf{e},\mathbf{q}) - \delta)} \Big) \ + \\
                   & \frac{1}{|\mathcal{Q}|} \sum_{\mathbf{q} \in \mathcal{Q}} \log \Big(1 + \sum_{\mathbf{e} \in \mathcal{E}_{\mathbf{q}}^{-}} e^{\alpha(\mathsf{sim}(\mathbf{e},\mathbf{q}) + \delta)} \Big),
\end{split}
$}
\label{eq:auxiliaryloss}
\end{equation}
where $\mathcal{Q} = \{ \mathbf{q}_1,...,\mathbf{q}_C \}$ denotes the set of learnable proxies for all classes (a proxy is a global representative class-wise anchor in the embedding space~\cite{kim2020proxy} and one proxy for each class), and 
$\mathcal{Q}^{+} \subset \mathcal{Q}$ is the set of proxies for the classes present in the mini-batch $\mathcal{K}$.
The set of embeddings $\mathbf{e}$ of the samples, computed with $\mathsf{GAP(\cdot)}$, from the mini-batch $\mathcal{K}$ is divided into the sets $\mathcal{E}_{\mathbf{q}}^{+}$ and $\mathcal{E}_{\mathbf{q}}^{-}$ containing the batch of embeddings with the same class or different class as the proxy $\mathbf{q}$, respectively. 
In Eq.~\eqref{eq:auxiliaryloss}, the function $\mathsf{sim}(\cdot,\cdot)$ computes the cosine similarity, 
$\delta$ and $\alpha$ are hyper-parameters ($\delta=0.1$ and $\alpha=32$, as in the original paper~\cite{kim2020proxy}).

As detailed in Algorithm~\ref{alg:training}, the main training procedure of MGProto alternates between two steps: 
1) optimising the model backbone and add-on layer $\left\{ \theta_{\textbf{bcb}}, \theta_{\textbf{add}} \right\}$ using Eq.~(\ref{eq:totalloss}), 
with the prototype distributions $\left\{\mathbf{p}_m^c, \boldsymbol{\pi}_m^c \right\}$ frozen; and 
2) estimating the prototype distributions via the modified EM algorithm through Eq.~(\ref{eq:e-step-closeform}), (\ref{eq:m_step_loss}), and (\ref{eq:m-step-pi}), 
with the model backbone and add-on layers frozen. 
Note that an additional warm-up stage is needed to fill the memory queue $\mathcal{B}^c$ with sufficient class-relevant features. 


\begin{algorithm}[!tb]
\caption{Training Procedure of MGProto}
\label{alg:training}

\KwData{Training set $\mathcal{D}$, Training epochs $E$.}

\KwResult{Model backbone $\theta_{\textbf{bcb}}$ and add-on layer $\theta_{\textbf{add}}$, \\ 
\qquad \quad \, Prototypes $\{\mathbf{p}_m^c, \boldsymbol{\pi}_m^c\}_{m=1}^{M}$ for $c \in \{1,...,C\}.$ 
}

\tcc{Alternate Training}
\For{$E$ Epochs}{
  Given a mini-batch $\mathcal{K}$ sampled from $\mathcal{D}$; \\
  Compute loss $\mathcal{L}_{total}$ in Eq.~(\ref{eq:totalloss}); \\
  Update backbone $\theta_{\textbf{bcb}}$ and add-on layer $\theta_{\textbf{add}}$; \\
  Update memory queue $\mathcal{B}^c$ for each class $c$; \\
  \For{$L_{em}$ Loops}{
      E-step: compute responsibility in Eq.~(\ref{eq:e-step-closeform}); \\
      M-step: update $\mathbf{p}_m^c$ in Eq.~(\ref{eq:m_step_loss}) and \\
\qquad \ \ \ \ \, update $\boldsymbol{\pi}_m^c$ in Eq.~(\ref{eq:m-step-pi}) ; \\
}}

\tcc{Prototype Replacement}
Ground prototype means $\mathbf{p}_m^c$ in Eq.~(\ref{eq:prototype_replacement_my}); 

\tcc{Model Pruning (Optional)}
Keep prototypes with top-$\widetilde{M}$ priors for each class $c$. 

\end{algorithm}

Although the optimisation of our MGProto includes a natural prototype replacement, 
we still need a way of visualising and grounding the prototypes in the image space, 
so that they are represented by actual training image patches.
Similar to Eq.~(\ref{eq:prototype_replacement}), we accomplish this by replacing each of the prototype means with the latent feature vector of its most active training image patch from the same class, 
with: 
\begin{equation}
\ \mathbf{p}_m^c \leftarrow \arg\max_{\mathbf{f} \in \mathbf{F}_{a\in\{1,...,|\mathcal{D}_c|\}}}  \mathbf{H}_m^{c},
\label{eq:prototype_replacement_my}
\end{equation}
where $\mathbf{H}_m^{c}$ is the likelihood map defined in Eq.~(\ref{eq:gaussian}). 
As shown in Fig.~\ref{fig:replacement}, our MGProto no longer suffers from performance degradation in the replacement step.

\subsection{MGProto Pruning by Prototype Importance Prior}

Pruning is a good way of reducing the explanation size of interpretable prototypical-part networks~\cite{doshi2017towards}. 
In practice, it is often implemented by discarding prototypes that are irrelevant, meaningless, or ambiguous~\cite{chen2019looks,rymarczyk2021protopshare,nauta2021neural,bontempelli2023concept}. 

In this work, we propose to remove less-important or trivial prototypes in a trained MGProto model, 
resorting to the inherent mixture constraint of GMMs: $\sum_{m = 1}^{M} \boldsymbol{\pi}_m^c = 1$ for each class $c$, as mentioned in Section~\ref{sec:mixtureofgaussian}. 
To be specific, we take advantage of the prototype importance prior $\boldsymbol{\pi}_m^c$, where
a large (or low) prior means the respective prototype is important (or trivial) to characterise the underlying class-conditional distribution $p(\mathbf{x}|c)$.
Based on the above analysis, we prune a trained MGProto by keeping only important prototypes corresponding to the top-$\widetilde{M}$ priors for each class, where $\widetilde{M} < M$.
This can avoid choosing complicated per-class pruning thresholds, which is required in previous pruning schemes~\cite{chen2019looks,nauta2021neural}.



\section{Experiments}

\subsection{Datasets and Evaluation Metrics}
We evaluate our method on three standard fine-grained image recognition benchmarks: CUB-200-2011~\cite{wah2011caltech}, Stanford Cars~\cite{krause20133d}, and Stanford Dogs~\cite{khosla2011novel}. 
To follow the same setting of OoD detection in PIP-Net~\cite{nauta2023pip}, we also use Oxford-IIIT Pets~\cite{parkhi2012cats}.
Images are resized to 224 $\times$ 224 and we apply the same online augmentations~\cite{nauta2021neural,ukai2023looks} to training images. 


Image recognition and OoD detection performances are evaluated using the top-1 accuracy and FPR95 metrics, respectively.
Following~\cite{nauta2023pip,huang2023evaluation,wang2023learning}, 
we quantify different facets of interpretability on CUB using full images, with: 
1) \textit{Consistency} score that quantifies how consistently each prototype activates the same human-annotated object part~\cite{huang2023evaluation}; 
2) \textit{Purity} of prototypes which, similarly to the consistency score, evaluates the extent that the top-10 image patches for a prototype can encode the same object-part~\cite{nauta2023pip}; 
3) \textit{Stability} score that measures how robust the activation of object parts is when noise is added to an image input~\cite{huang2023evaluation};
4) \textit{Outside-Inside Relevance Ratio} (OIRR) which calculates the ratio of mean activation outside the object to those within the object, using ground-truth object segmentation masks~\cite{lapuschkin2016analyzing}; and 5) \textit{Deletion AUC} (DAUC) that computes the degree in the probability drop of the predicted class as more and more activated pixels are erased~\cite{petsiuk2018rise}.
The above consistency, purity, and stability are part-level measures, OIRR is an object-level measure, and DAUC is a model-level measure based on causality.

\subsection{Implementation Details}

We perform experiments on various CNN backbone architectures: VGG16 (V16), VGG19 (V19), ResNet34 (R34), ResNet50 (R50), ResNet152 (R152), DenseNet121 (D121), and DenseNet161 (D161), which are all pre-trained on ImageNet~\cite{deng2009imagenet}, except for ResNet50 on CUB that is pre-trained on iNaturalist~\cite{van2018inaturalist}. 
In accordance with prior studies, the prototype dimension $D=64$ and prototype number $M=10$ for all backbones and datasets. 
Following~\cite{donnelly2022deformable,nauta2023pip}, we obtain more fine-grained feature maps with $\bar{H} = \bar{W} = 14$ by dropping the final max-pooling layer in the backbones. 
Then two add-on $1 \times 1$ convolutional layers (without activation function\footnote{
Two convolutional layers without a non-linear activation between them are functionally similar to applying a single convolutional layer. 
Here we choose the two-layer structure for consistency with many prior ProtoPNet-based methods.}
) are appended to reduce the number of feature channels to match the prototype dimension, thereby improving the computational efficiency.
The memory capacity $N$ is set to 800, 1000, 2000, 2000 for CUB, Cars, Dogs, and Pets, respectively. 
In Eq.~(\ref{eq:totalloss}), we have $\lambda_1$ = 0.2, and $\lambda_2$ is set as 0.5 according to~\cite{ukai2023looks}. 
For the prototype mining, we use $T$ = 20 levels of active patches.

Our MGProto is implemented with PyTorch~\cite{paszke2019pytorch} and trained with Adam optimiser, using a mini-batch size $|\mathcal{K}| = 80$.
The learning rates for the CNN backbone are chosen as $1 \times 10^{-4}$ (CUB, Cars, and Pets) and $1 \times 10^{-5}$ (Dogs). 
The learning rates of add-on layers are set as $3 \times 10^{-3}$ for all datasets.
These learning rates are decreased by 0.4 every 15 epochs (with a total of $E$ = 120 training epochs). 
For the optimisation of prototype means in Eq.~(\ref{eq:m_step_loss}), the learning rates of gradient descent are $3 \times 10^{-3}$ (CUB) and $3 \times 10^{-4}$ (Cars, Dogs, and Pets). 
We set the number of EM loops $L_{em}$ = 3.
To improve the EM solution, an exponential moving average (EMA)~\cite{hunter1986exponentially} is used in the M-step: 
$\{\boldsymbol{\pi}_m^c\}^{t+1}_{\rm ema} := \tau \{\boldsymbol{\pi}_m^c\}^{t}_{\rm ema} + (1 - \tau) \{\boldsymbol{\pi}_m^c\}^{t}$, where $\tau = 0.99$. 
Also, we experimentally noticed that for a few classes, all feature samples in the memory queue tend to have a large responsibility for a single Gaussian prototype in the E-step, ignoring other prototypes. 
We alleviate this issue with an additive smoothing~\cite{chen1999empirical,valcarce2016additive} to soften the calculated responsibility in Eq.~(\ref{eq:e-step-closeform}) for each feature sample: 
$\gamma_{n, m}^{c,\mathsf{soft}} = \nicefrac{\left(\gamma_{n, m}^c + \alpha\right)}{\left(\sum_{m=1}^{M}(\gamma_{n, m}^c + \alpha)\right)}$, where $\alpha = 0.1$. 


\subsection{Comparison With SOTA Methods}
\subsubsection{Accuracy of Interpretable Image Recognition}

\begin{table}[]
\setlength{\tabcolsep}{0.3 mm}
\caption{Classification accuracy (\%) on full images of CUB-200-2011, where 1×1p, 10pc denotes 1$\times$1 prototype shape and 10 prototypes per class. 
We report our results as the mean and standard deviation over 5 runs.
}
\vspace{-9pt}
\resizebox{\linewidth}{!}{

\begin{tabular}{l|cccccccc} 
\hline
Method                                                 & \# Proto.       & V16         & V19         & R34         & R50         & R152      & D121       & D161  \\ 
\hline
Baseline                                               & n.a.            & 70.9        & 71.3        & 76.0        & 78.7        & 79.2      & 78.2        & 80.0      \\ 
\hline
ProtoPool \cite{rymarczyk2022interpretable}            & 1×1p, 202       & 74.8        & 75.3        & 76.2        & 83.4        & 79.9      & 78.1        & 80.5      \\
ProtoKNN \cite{ukai2023looks}                          & 1×1p, 512       & 77.2        & 77.6        & 77.6        & 87.0        & 80.6      & 79.8        & 81.4      \\
MGProto          & 1×1p, 400       & 79.1\tiny{(0.3)}     & 79.3\tiny{(0.1)}     & 81.0\tiny{(0.2)}    & 87.3\tiny{(0.2)}    & 81.6\tiny{(0.1)}     & 80.6\tiny{(0.2)}    & 82.3\tiny{(0.3)}    \\ 
\hline
ProtoPNet \cite{chen2019looks}                         & 1×1p, 10pc      & 70.3        & 72.6        & 72.4        & 81.1        & 74.3      & 74.0        & 75.4      \\
ProtoConcepts \cite{ma2024looks}                       & 1×1p, 10pc      & 70.6        & 72.7        & 73.0        & 81.7        & 74.7      & 73.8        & 75.7      \\
TesNet \cite{wang2021interpretable}                    & 1×1p, 10pc      & 75.8        & 77.5        & 76.2        & 86.5        & 79.0      & 80.2        & 79.6      \\
Deformable \cite{donnelly2022deformable}               & 2×2p, 10pc      & 75.7        & 76.0        & 76.8        & 86.4        & 79.6      & 79.0        & 81.2      \\
ST-ProtoPNet \cite{wang2023learning}                   & 1×1p, 10pc      & 76.2        & 77.6        & 77.4        & 86.6        & 78.7      & 78.6        & 80.6      \\
SDFA-SA \cite{huang2023evaluation}                    & 1×1p, 10pc       & 76.4        & 77.7        & 77.8        & 86.4        & 79.9      & 80.4        & 81.4      \\
MGProto           & 1×1p, 10pc      & \textbf{80.6}\tiny{(0.3)} & \textbf{80.4}\tiny{(0.1)} & \textbf{82.2}\tiny{(0.2)}  & \textbf{87.8}\tiny{(0.1)}  & \textbf{82.8}\tiny{(0.2)} & \textbf{81.8}\tiny{(0.3)} & \textbf{84.1}\tiny{(0.2)}  \\
\hline
\end{tabular}

}
\label{tab:fullCUB}
\end{table}

Table~\ref{tab:fullCUB} shows the comparison results with other SOTA ProtoPNet variants using full CUB images, where the Baseline denotes a non-interpretable black-box model. 
Across all CNN backbones, our MGProto consistently achieves the highest accuracy, when using 10 prototypes per class. 
Even when trained with fewer prototypes (2 per class, totalling 400), MGProto remains highly effective, outperforming other methods like ProtoPool and ProtoKNN by a significant margin.
Table~\ref{tab:Cars} shows the classification results on Cars. 
Following~\cite{chen2019looks,rymarczyk2022interpretable,ukai2023looks}, 
we also crop Car images using the provided bounding boxes to perform experiments. 
As evident, our MGProto exhibits the best accuracy under both settings of full and cropped images. 
Table~\ref{tab:fullDogs} further displays the superiority of MGProto on Stanford Dogs.

\begin{figure*}[t!]
    \centering
    \includegraphics[width=1.00\linewidth]{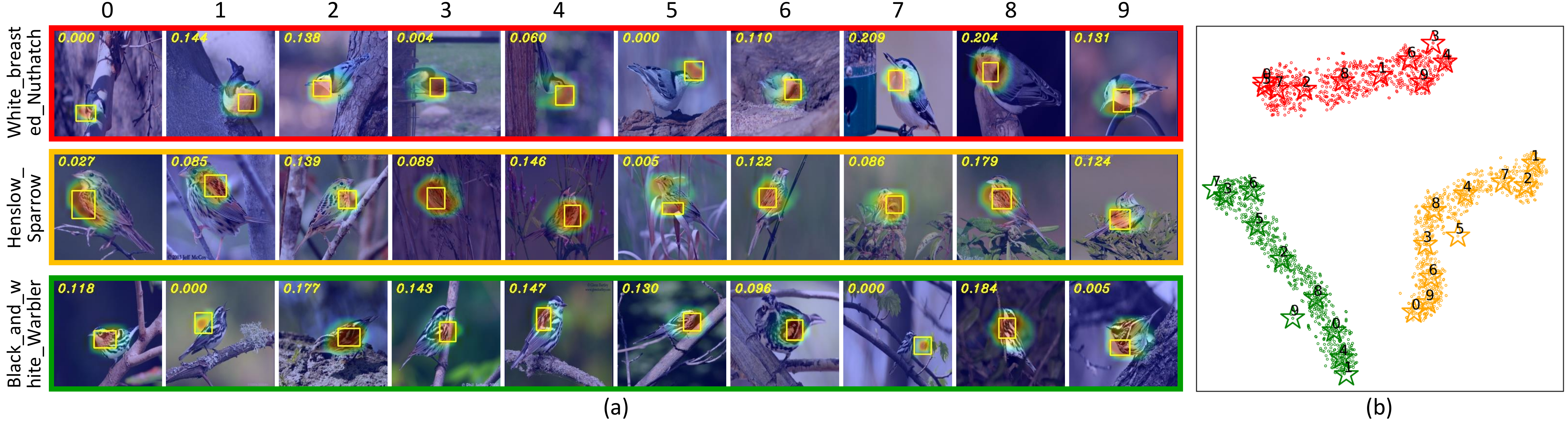}
    \vspace{-23pt}
    \caption{Visual prototypes (a) and T-SNE results (b) from a ResNet34-based MGProto model, trained on CUB. 
    These prototypes are marked with a yellow box in the activation maps, with the corresponding importance prior given in yellow text. 
    In T-SNE, we show the prototypes (stars), their ID (numbers), and the nearest training features (dots) stored in the memory queue. 
    This figure randomly shows 3 out of 200 classes, where each colour indicates a different class and for each class we display all 10 prototypes.
    }
    \label{fig:prototype}
    \vspace{-5pt}
\end{figure*}

Fig.~\ref{fig:prototype} visualises all 10 prototypes for 3 classes from CUB, and T-SNE representations of prototypes and training features from our MGProto. 
It can be observed that: 
1) large-prior prototypes, which dominate the decision making, are always from high-density distribution regions (in T-SNE) and can localise well the object (bird) parts;
2) background prototypes tend to have a low prior and come from the low-density distribution regions;
3) some prototypes localise the object parts but have a low prior, suggesting that those object parts may not be important for identifying that class; 
4) the prototypes naturally localise at the centre of data clusters (in T-SNE), which prevents a decrease in performance when replacing prototypes with the nearest features.


\begin{table}[]
\setlength{\tabcolsep}{0.25 mm}
\caption{
Classification accuracy (\%) on Stanford Cars based on ResNet50 backbone, trained and tested using "Crop" and "Full" car images. We report our results as the mean and standard deviation over 5 runs.
}
\vspace{-9pt}
\resizebox{\linewidth}{!}{

\begin{tabular}{lcccccc} 
\hline
Data          & ProtoPNet~\cite{chen2019looks} & ProtoTree~\cite{nauta2021neural} & ProtoPool~\cite{rymarczyk2022interpretable} & PIP-Net~\cite{nauta2023pip} & ProtoKNN~\cite{ukai2023looks} & MGProto  \\ 
\hline
Crop        & 88.4      & 89.2      & 88.9      & 90.2       & 90.9     & \textbf{92.0}\tiny{(0.1)}      \\
Full        & 86.1      & 86.6      & 86.3      & 86.5       & –        & \textbf{89.2}\tiny{(0.2)}      \\
\hline
\end{tabular}

}
\label{tab:Cars}
\end{table}

\begin{table}[]
\setlength{\tabcolsep}{0.6 mm}
\caption{Classification accuracy (\%) on full images of Stanford Dogs. We report our results as the mean and standard deviation over 5 runs.
}
\vspace{-9pt}
\resizebox{\linewidth}{!}{

\begin{tabular}{l|cccccccc} 
\hline
Method                                                         & \# Proto.     & V16         & V19       & R34         & R50         & R152      & D121       & D161     \\ 
\hline
Baseline                                                       & n.a.          & 75.6        & 77.3      & 81.1        & 83.1        & 85.2        & 81.9        & 84.1         \\ 
\hline
ProtoPNet \cite{chen2019looks}                                 & 1×1p, 10pc    & 70.7        & 73.6      & 73.4        & 76.4        & 76.2        & 72.0        & 77.3         \\
ProtoConcepts \cite{ma2024looks}                               & 1×1p, 10pc    & 71.2        & 73.8      & 73.7        & 76.2        & 76.9        & 72.5        & 77.5         \\
TesNet \cite{wang2021interpretable}                            & 1×1p, 10pc    & 74.3        & 77.1      & 80.1        & 82.4        & 83.8        & 80.3        & 83.8         \\
Deformable \cite{donnelly2022deformable}                       & 3×3p, 10pc    & \textbf{75.8}  & 77.9      & 80.6        & 82.2        & 86.5        & 80.7        & 83.7         \\
ST-ProtoPNet \cite{wang2023learning}                           & 1×1p, 10pc    & 74.2        & 77.2      & 80.8        & 84.0        & 85.3        & 79.4        & 84.4         \\
SDFA-SA \cite{huang2023evaluation}                             & 1×1p, 10pc    & 73.9        & 77.1      & 81.4        & 84.3        & 85.9        & 80.1        & 84.7         \\
MGProto                 & 1×1p, 10pc    & 75.6\tiny{(0.2)}    & \textbf{78.8}\tiny{(0.1)}    & \textbf{83.6}\tiny{(0.3)}  & \textbf{85.9}\tiny{(0.2)}  & \textbf{87.7}\tiny{(0.1)}   & \textbf{81.9}\tiny{(0.3)}   & \textbf{86.1}\tiny{(0.3)}    \\
\hline
\end{tabular}

}
\label{tab:fullDogs}
\end{table}

\begin{table}[]
\setlength{\tabcolsep}{0.8 mm}
\caption{FPR95 (\%) of OoD detection for each ID-OoD pair, computing the false positive rate (FPR) of OoD samples when the true positive rate of ID samples is at 95\%. The models are trained using only ID samples. 
}
\vspace{-9pt}
\resizebox{\linewidth}{!}{

\begin{tabular}{lcccccc} 
\hline
Method                                      & CUB-Cars         & CUB-Pets        & Cars-CUB        & Cars-Pets        & Pets-CUB &        Pets-Cars             \\ 
\hline
PIP-Net~\cite{nauta2023pip}                 & 1.1              & 8.1             & 7.8             & 5.6              & 12.9              & 0.9               \\
MGProto                                     & \textbf{0.05}  & \textbf{7.6}  & \textbf{3.2}  & \textbf{2.0}   & \textbf{8.0}    & \textbf{0.08}     \\
\hline
\end{tabular}

}
\label{tab:OoD}
\end{table}

\subsubsection{Accuracy of Trustworthy Image Recognition}

\begin{figure}[t!]
    \centering
    \includegraphics[width=0.8\linewidth]{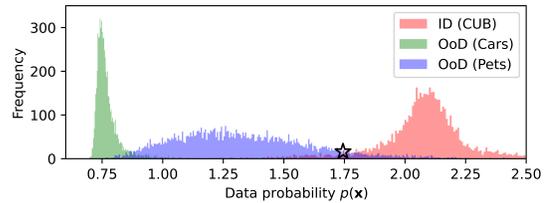}
    \vspace{-12pt}
    \caption{Histograms of the overall data probability $p(\mathbf{x})$ computed on ID's test set and OoD's datasets. 
    MGProto abstains from making a decision on an input $\mathbf{x}$ if $p(\mathbf{x})$ falls below a threshold marked by the star. 
    }
    \label{fig:histogram}
\end{figure}

Our MGProto is capable of identifying OoD inputs and abstaining from making decisions on them, thereby achieving trustworthy image recognition, through Eq.~(\ref{eq:ood}). 
Following PIP-Net~\cite{nauta2023pip}, we use ResNet50 backbone and perform experiments on CUB, Cars, and Pets (e.g., choose one dataset as ID and the other two as OoD). 
Table~\ref{tab:OoD} shows that MGProto outperforms PIP-Net by a large margin in most tasks. 
For example, when trained on Cars  and classifying 95\% of the testing Cars images as ID samples, MGProto detects 96.8\% of CUB images as OoD, while PIP-Net only detects 92.2\% under the same setting. 
Fig.~\ref{fig:histogram} depicts the probability histogram for the task of CUB as ID, Cars and Pets as OoD, 
revealing that our MGProto produces high probability for ID samples and low probability for OoD samples.
Fig.~\ref{fig:histogram} also suggests that CUB images have smaller semantic distance to Pets than Cars, which is aligned with our human intuition. 

\begin{figure*}[t!]
    \centering
    \includegraphics[width=1.0\linewidth]{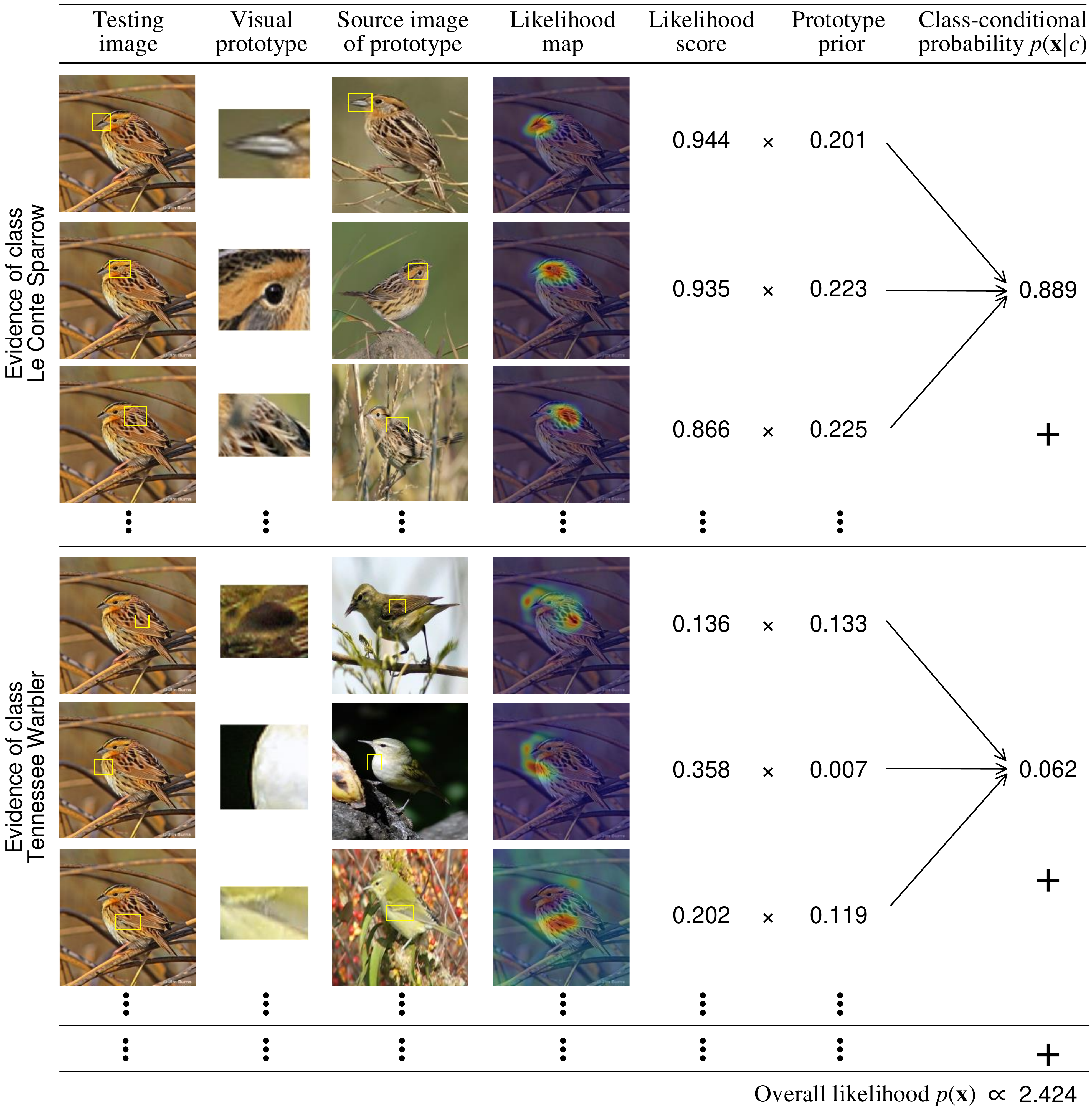}
    \vspace{-24pt}
    \caption{Interpretable and trustworthy reasoning of our MGProto method for (a) classifying an In-Distribution Le Conte Sparrow image from CUB test set and (b) identifying an Out-of-Distribution Keeshond image from Pets. In the figure, we only show the top-2 classification predictions.}
    \label{fig:reasoning}
\end{figure*}

Fig.~\ref{fig:reasoning}(a) displays an example of the interpretable and trustworthy reasoning of our MGProto for classifying an ID sample (Le Conte Sparrow image from CUB test set). 
In this case, the model obtains the highest $p(\mathbf{x}|c)$ for the class Le Conte Sparrow among all classes, 
meaning that the input image data fits best the prototype distributions of this class. 
Hence, our model classifies the image as Le Conte Sparrow. 
Additionally, our method evaluates the overall data probability $p(\mathbf{x})$ that the input data belongs to the distribution of the training set.
Here, the $p(\mathbf{x}) \propto 2.429$, surpassing the OoD detection threshold in Fig.~\ref{fig:histogram}, 
our model thus treats the input image as an ID sample, which means that the classification decision is trustworthy and should be accepted. 
Fig.~\ref{fig:reasoning}(b) shows an example of our MGProto method recognising an OoD sample (Keeshond image from Pets), 
where the input image obtains $p(\mathbf{x})$ smaller than the OoD detection threshold. 
The model opts to abstain from classifying the image since it does not fit well the CUB-prototype distributions of all 200 classes.

Notice that the OoD detection is a long-standing research topic, and mainstream approaches harness post-hoc analysis, 
which builds a scoring criterion (e.g., logit-based~\cite{basart2022scaling,liu2020energy}, distance-based~\cite{lee2018simple,sun2022out}, and density-based~\cite{morteza2022provable,peng2024conjnorm}) to indicate the ID-ness of the input. 
Relying on these scores, subsequent studies boost the OoD detection with auxiliary training data from outlier exposure~\cite{hendrycks2019deep} or synthesis~\cite{tao2023non}. 
To explore where the ProtoPNet-based methods stand in comparison with specialised OoD detection approaches,
we also experiment on the standard CIFAR-10 benchmark. 
Consistent with the setup in~\cite{sun2022out,liu2024neuron}, we utilise CIFAR-10~\cite{krizhevsky2009learning} as ID dataset and the same model backbone (ResNet-18) in our method. 
OoD datasets include: 
SVHN~\cite{netzer2011reading}, Places365~\cite{zhou2017places}, LSUN~\cite{yu2015lsun}, iSUN~\cite{xu2015turkergaze}, and Textures~\cite{cimpoi2014describing}.
In line with prior post-hoc approaches, we employ only ID data for model's training, without using any auxiliary OoD training data. 
Experimental results in Table~\ref{tab:OoD_more} demonstrate that our distribution-based prototypes achieve the best results on Textures dataset and deliver comparable average performances on par with other leading OoD detection approaches. 
In particular, our MGProto method greatly outperforms GEM~\cite{morteza2022provable} that models the class-conditional density using only a single Gaussian distribution, 
since our method leverages mixture of Gaussian distributions, enabling a more accurate characterization of the true underlying data density for each class. 

\begin{table*}[]
\setlength{\tabcolsep}{2.1 mm}
\centering
\caption{Comparison with specialised OoD detection approaches, evaluated with FPR95 (\%) and AUROC (\%). We show the results on each individual OoD dataset and  the results on average across all  five OoD datasets.}
\vspace{-9pt}
\resizebox{\linewidth}{!}{

\begin{tabular}{lcccccccccccc} 
\hline
\multirow{3}{*}{Method} & \multicolumn{12}{c}{OoD Dataset}                                                                                                                                             \\
                        & \multicolumn{2}{c}{SVHN} & \multicolumn{2}{c}{Places365} & \multicolumn{2}{c}{LSUN} & \multicolumn{2}{c}{iSUN} & \multicolumn{2}{c}{Textures} & \multicolumn{2}{c}{Average}  \\
                        & FPR95↓ & AUROC↑          & FPR95↓ & AUROC↑               & FPR95↓ & AUROC↑          & FPR95↓ & AUROC↑          & FPR95↓ & AUROC↑              & FPR95↓ & AUROC↑              \\ 
\hline
MSP~\cite{hendrycks2017baseline}       & 59.66  & 91.25           & 62.46  & 88.64               & 51.93  & 92.73           & 54.57  & 92.12          & 66.45  & 88.50           & 59.01  & 90.65   \\
ODIN~\cite{liang2018enhancing}         & 20.93  & 95.55           & 63.04  & 86.57               & 31.92  & 94.82           & 33.17  & 94.65          & 56.40  & 86.21           & 41.09  & 91.56   \\
Energy~\cite{liu2020energy}            & 54.41  & 91.22           & 42.77  & 91.02               & 23.45  & 96.14           & 27.52  & 95.59          & 55.23  & 89.37           & 40.68  & 92.67   \\
Mahalanobis~\cite{lee2018simple}  & \textbf{9.24}   & \textbf{97.80}    & 83.50  & 69.56         & 67.73  & 73.61  & \textbf{6.02}   & \textbf{98.63} & 23.21  & 92.91           & 37.94  & 86.50   \\
ViM~\cite{wang2022vim}                 & 24.95  & 95.36           & 63.04  & 86.57   & \textbf{7.26}   & \textbf{98.53}     & 33.17  & 94.65          & 56.40  & 86.21           & 36.96  & 92.26   \\
ReAct~\cite{sun2021react}              & 48.16  & 92.32           & 37.25  & 93.13               & 18.09  & 96.91           & 20.35  & 95.59          & 96.51  & 47.41           & 34.25  & 94.09   \\
KNN~\cite{sun2022out}                  & 24.53  & 95.96           & 50.90  & 89.14               & 25.29  & 95.69           & 25.55  & 95.26          & 27.57  & 94.71           & 30.77  & 94.15   \\
ConjNorm~\cite{peng2024conjnorm}       & 18.71  & 96.48           & 53.44  & 89.18               & 22.20  & 95.95           & 22.64  & 95.87          & 25.80  & 95.11           & 28.56  & 94.52   \\
ASH~\cite{djurisic2023extremely}       & 28.94  & 94.84           & 27.29  & 91.31               & 9.06   & 98.34           & 21.61  & 95.95          & 35.02  & 93.63           & 27.29  & 94.81   \\
GEM~\cite{morteza2022provable}         & 37.41  & 93.22           & 38.40  & 91.50               & 10.81  & 97.90           & 12.36  & 97.57          & 28.88  & 94.38           & 25.57  & 94.91   \\
GEN~\cite{liu2023gen}                  & 26.10  & 93.83           & 37.40  & \textbf{93.42}      & 8.99   & 98.39           & 24.09  & 95.03          & 26.62  & 94.13           & 24.64  & 94.96   \\
NCA~\cite{liu2024neuron}               & 14.33  & 96.05  & \textbf{26.73}  & 91.85               & 19.28  & 96.83           & 29.88  & 94.86          & 17.03  & 95.64  & \textbf{21.45}  & 95.05   \\
MGProto                                & 28.56  & 94.38           & 39.40  & 91.38               & 10.74  & 98.03    & 28.20  & 94.75   & \textbf{12.26}  & \textbf{97.59}   & 23.83  & \textbf{95.23} \\
\hline
\end{tabular}

}
\label{tab:OoD_more}
\end{table*}


\subsubsection{Pruning of MGProto Model}
\label{se:prototype_prunning}

Table~\ref{tab:pruneCUB} shows the MGProto pruning results. 
We first attempt to apply the purity-based pruning strategy~\cite{chen2019looks} to our MGProto, 
but only a few prototypes are removed. 
This means that the strategy is improper to our method, mostly because our prototypes have large purity, evidenced in Table~\ref{tab:purity}. 
On the other hand, if the merge-based pruning~\cite{rymarczyk2021protopshare} is adopted,
the model's performance will be heavily sacrificed despite a substantial number of pruned prototypes. 

Fortunately, by using only top-$\widetilde{M}$ important ($\widetilde{M}$ = 1, 2, 4, 6, 8) prototypes per class, our importance-based strategy can greatly prune MGProto, while maintaining a high classification accuracy. 
Interestingly, discarding 20\% of prototypes slightly leads to increased accuracy. 
With a very large pruning rate (e.g., 90\%), our method exhibits a tolerable performance decrease (about 2\% in both backbones). 
This is because the large-prior prototypes always lie in high-density regions of the data distribution (see Fig.~\ref{fig:prototype}(b)), 
capturing enough representative and crucial information.
It could be imagined that removing those small-prior prototypes does not render significant changes in the decision boundary.

\begin{table}[]
\setlength{\tabcolsep}{0.4 mm}
\caption{Model pruning results of our MGProto on CUB-200-2011, using different pruning strategies.
\% Pruned denotes the ratio between the number of pruned prototypes to that of the initial ones.
}
\vspace{-9pt}
\resizebox{\linewidth}{!}{

\begin{tabular}{l|c|ccc|ccc} 
\hline
\multirow{2}{*}{Pruning strategy}         & \multirow{2}{*}{\begin{tabular}[c]{@{}c@{}}\# Initial~\\Proto.\end{tabular}} & \multicolumn{3}{c|}{ResNet34}              & \multicolumn{3}{c}{ResNet152}             \\
                                &                          & \# Proto.  & \% Pruned  & Acc. (\%)                & \# Proto.  & \% Pruned  & Acc. (\%)      \\ 
\hline
Purity-based~\cite{chen2019looks}                          & 2000                                      & 1993    & 0.35    & 82.2 $\rightarrow$ 82.2            & 1987   & 0.65    & 82.8 $\rightarrow$ 82.8        \\
Merge-based~\cite{rymarczyk2021protopshare}                & 2000                                      & 583     & 70.9    & 82.2 $\rightarrow$ 78.8            & 982    & 50.9    & 82.8 $\rightarrow$ 73.0        \\      
\hline
\multirow{5}{*}{\begin{tabular}[c]{@{}c@{}}Importance-based\\(ours)\end{tabular}}    & 2000            & 1600    & 20.0    & 82.2 $\rightarrow$ 82.4            & 1600   & 20.0    & 82.8 $\rightarrow$ 82.9        \\
                                                           & 2000                                      & 1200    & 40.0    & 82.2 $\rightarrow$ 82.0            & 1200   & 40.0    & 82.8 $\rightarrow$ 82.4        \\
                                                           & 2000                                      & 800     & 60.0    & 82.2 $\rightarrow$ 81.8            & 800    & 60.0    & 82.8 $\rightarrow$ 82.0        \\
                                                           & 2000                                      & 400     & 80.0    & 82.2 $\rightarrow$ 81.1            & 400    & 80.0    & 82.8 $\rightarrow$ 81.3        \\
                                                           & 2000                                      & 200     & 90.0    & 82.2 $\rightarrow$ 80.5            & 200    & 90.0    & 82.8 $\rightarrow$ 80.7        \\
\hline
\end{tabular}

}
\label{tab:pruneCUB}
\end{table}

\subsubsection{Interpretability Quantification}

Table~\ref{tab:interpretability} shows the interpretability results evaluated on CUB.
MGProto has the highest consistency score, indicating that the object parts activated by our prototypes are consistent across different images.
This is likely because our prototypes are holistically learned from a weighted average of a large set of relevant features, 
allowing them to consistently capture cross-image semantics. 
MGProto also achieves competitive stability score. 
The lowest OIRR score demonstrates that MGProto relies more on the object region and less on the background context to support its classification decision. 
The DAUC result indicates that MGProto produces interpretations that best influence its classification predictions. 
We also compute the prototype purity in Table~\ref{tab:purity}, showing that the prototypes produced by our MGProto have a high degree of purity, i.e., each prototype focuses on semantically-consistent object parts among different images. 

\begin{table}[]
\setlength{\tabcolsep}{0.50 mm}
\caption{Quantitative interpretability results (\%) on CUB test set using full images, where all models are based on VGG19 backbone. 
}
\vspace{-9pt}
\resizebox{\linewidth}{!}{

\begin{tabular}{lcccccc} 
\hline
Metric                  & \begin{tabular}[c]{@{}c@{}}ProtoPNet\\~\cite{chen2019looks}~\end{tabular} & \begin{tabular}[c]{@{}c@{}}Deformable\\~\cite{donnelly2022deformable}~\end{tabular} & \begin{tabular}[c]{@{}c@{}}TesNet\\~\cite{wang2021interpretable}~\end{tabular}  & \begin{tabular}[c]{@{}c@{}}ST-ProtoPNet\\~\cite{wang2023learning}~\end{tabular}  & \begin{tabular}[c]{@{}c@{}}SDFA-SA\\~\cite{huang2023evaluation}~\end{tabular}  & \begin{tabular}[c]{@{}c@{}}MGProto\\(ours)\end{tabular}        \\ 
\hline
Consistency ($\uparrow$)     & 45.29     & 57.87    & 60.75            & 74.08     & 80.45             & \textbf{93.21}\\
Stability ($\uparrow$)       & 42.23     & 43.91    & 39.20            & 44.96     & 46.30             & \textbf{46.95}    \\
OIRR ($\downarrow$)          & 37.26     & 28.68    & 38.97            & 28.09     & 33.65             & \textbf{22.30}\\
DAUC ($\downarrow$)          & 7.39      & 5.99     & 5.86             & 5.74      & 4.30              & \textbf{3.11}\\
\hline
\end{tabular}

}
\label{tab:interpretability}
\end{table}

\begin{table}[]
\centering
\setlength{\tabcolsep}{0.90 mm}
\caption{Prototype purity (\%) on train and test sets of CUB using full images, where all models are based on ResNet50 backbone. 
}
\vspace{-9pt}
\resizebox{0.62\linewidth}{!}{

\begin{tabular}{lcc} 
\hline
Method                                            & Purity (train) $\uparrow$         & Purity (test) $\uparrow$  \\ 
\hline
ProtoPNet~\cite{chen2019looks}                    & 0.44 ± 0.21            & 0.46 ± 0.22    \\
ProtoTree~\cite{nauta2021neural}                  & 0.13 ± 0.14            & 0.14 ± 0.16    \\
ProtoPShare~\cite{rymarczyk2021protopshare}       & 0.43 ± 0.21            & 0.43 ± 0.22    \\
ProtoPool~\cite{rymarczyk2022interpretable}       & 0.35 ± 0.20            & 0.36 ± 0.21    \\
PIP-Net~\cite{nauta2023pip}                       & 0.63 ± 0.25            & 0.65 ± 0.25    \\
MGProto                                           & \textbf{0.68} ± 0.22   & \textbf{0.69} ± 0.22    \\
\hline
\end{tabular}

}
\label{tab:purity}
\end{table}

\subsection{Ablation Studies}
\label{sec:ablation}

\subsubsection{Auxiliary Loss and Prototype Mining}
In Table~\ref{tab:ablationLoss}, we provide extensive ablation experiments on the auxiliary loss and prototype mining. 
Firstly, with the use of auxiliary loss $\mathcal{L}_{aux}$, the classification accuracy is improved, 
affirming its ability to aid the model backbone in enhancing feature extraction, as in~\cite{ukai2023looks}. 
However, $\mathcal{L}_{aux}$ does not have  much effect on the interpretability measures, 
because it is designed primarily for feature extraction rather than interpretability.

\begin{figure}[t!]
    \centering
    \includegraphics[width=1.0\linewidth]{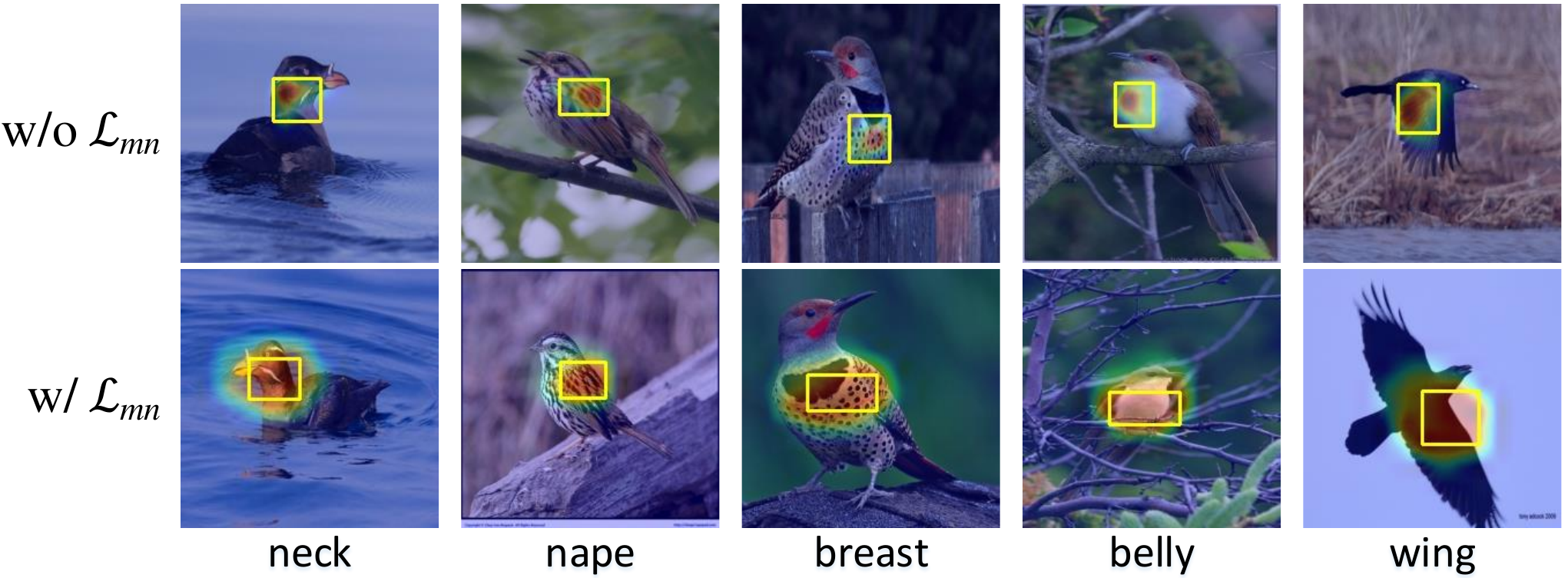}
    \vspace{-23pt}
    \caption{Visual comparison of prototypes learned without (w/o) and with (w/) prototype mining $\mathcal{L}_{mn}$, where each column shows prototypes from the same class, representing the same object part. 
    These boxes are all drawn around the largest 5\% of values of the activation map. }
    \label{fig:prototype_compare}
    \vspace{5pt}
\end{figure}

On the other hand, our prototype mining $\mathcal{L}_{mn}$ improves the accuracy 
since it explicitly enforces the prototypes of MGProto to capture more difficult-to-learn visual features from sub-salient object parts. 
Additionally, $\mathcal{L}_{mn}$ also enhances interpretability, notably regarding OIRR and DAUC, 
because $\mathcal{L}_{mn}$ enforces prototypes to activate more sub-salient object regions, as shown in Fig.~\ref{fig:prototype_compare}. 
We notice a marginal improvement of consistency score compared to other metrics. 
We argue that this is because the prototype learning using $\mathcal{L}_{ce}$ can already acquire highly consistent prototypes across images,
by holistically leveraging a large set of relevant features of different samples.

To explore the generality of our horse-racing based prototype mining strategy, 
we also apply it to other existing prototypical-part methods, e.g., ProtoPNet, TesNet, and ST-ProtoPNet.
Results in Table~\ref{tab:generality} show that our prototype mining strategy is still effective in these existing models relying on point-based learning techniques. 
We notice the accuracy improvement is particularly significant in the vanilla ProtoPNet~\cite{chen2019looks},
mostly because this baseline model has the worse classification performance among the three models.

\begin{table}[]
\centering
\setlength{\tabcolsep}{0.6 mm}
\caption{Ablation analysis of the compared
mining loss $\mathcal{L}_{mn}$ in Eq.~\eqref{eq:mining} and auxiliary loss $\mathcal{L}_{aux}$ in Eq.~\eqref{eq:auxiliaryloss} for the proposed MGProto method. 
}
\vspace{-9pt}
\resizebox{1.0\linewidth}{!}{

\begin{tabular}{ccc|ccc|cccc}
\hline
\multirow{2}{*}{$\mathcal{L}_{ce}$} & \multirow{2}{*}{$\mathcal{L}_{aux}$} & \multirow{2}{*}{$\mathcal{L}_{mn}$} & \multicolumn{3}{c|}{Accuracy (\%)} & \multicolumn{4}{c}{Interpretability (\%), V19}  \\
                                    &                                     &  & CUB(R34) & Dogs(R34) & Cars(R50)   & Consist.($\uparrow$) & Stability($\uparrow$) & OIRR($\downarrow$) & DAUC($\downarrow$)      \\ 
\hline
$\checkmark$                        &                                     &                                      & 79.6     & 81.0      & 87.0       & 92.97& 45.67& 23.83& 3.91\\
$\checkmark$                        & $\checkmark$                                    &                          & 81.1     & 82.5      & 88.7       & 92.90& 46.06   & 23.70   & 3.52\\
$\checkmark$                        &                                     & $\checkmark$                         & 81.5     & 82.9      & 88.5       & 93.11& 46.38& 22.33& 3.20\\
$\checkmark$                        & $\checkmark$                        & $\checkmark$                         & 82.2     & 83.6      & 89.2       & 93.21& 46.95   & 22.30& 3.11\\
\hline
\end{tabular}

}
\label{tab:ablationLoss}
\end{table}

\begin{table}[]
\setlength{\tabcolsep}{2.2 mm}
\caption{Improved accuracy (\%) of existing prototypical-part networks on CUB, integrated with our prototype mining, based on ResNet34 backbone.
}
\vspace{-9pt}
\resizebox{\linewidth}{!}{

\begin{tabular}{cc|cc|cc}
\hline
\multicolumn{2}{c|}{ProtoPNet} & \multicolumn{2}{c|}{TesNet} & \multicolumn{2}{c}{ST-ProtoPNet}  \\ 
\hline
Vanilla~\cite{chen2019looks}  & + Mining               & Vanilla~\cite{wang2021interpretable}  & + Mining                  & Vanilla~\cite{wang2023interpretable}   & + Mining   \\ 
72.4                          & 76.5                   & 76.2                                  & 78.6                      & 77.4                                   & 79.3       \\
\hline
\end{tabular}

}
\label{tab:generality}
\end{table}

\subsubsection{Prototype Diversity in M-step} 
In Eq.~(\ref{eq:m_step_loss}), we introduce a diversity constraint to improve the within-class distances between prototype means.   
Table~\ref{tab:diversity} presents an ablation study using full images of CUB and Cars.
Our approach in Eq.~(\ref{eq:m_step_loss}) can significantly increase the average pair-wise distances ($L_2$) between prototype means. 
Also, prototypes with greater diversity tend to correlate with higher classification accuracy, 
in line with findings observed in established methods~\cite{wang2021interpretable,donnelly2022deformable,wang2022knowledge}. 

\begin{table}[]
\setlength{\tabcolsep}{2.6 mm}
\caption{Effect of the prototype diversity constraint in Eq.~(\ref{eq:m_step_loss}) on the within-class distance between prototype means and classification accuracy (\%).
}
\vspace{-9pt}
\resizebox{\linewidth}{!}{

\begin{tabular}{cc|cc|cc|cc} 
\hline
\multicolumn{4}{c|}{CUB(R34)}                                                & \multicolumn{4}{c}{Cars(R50)}                      \\ 
\hline
\multicolumn{2}{c|}{Closed-form Eq.~(\ref{eq:m-step-closeform})} & \multicolumn{2}{c|}{Our Eq.~(\ref{eq:m_step_loss})} & \multicolumn{2}{c|}{Closed-form Eq.~(\ref{eq:m-step-closeform})} & \multicolumn{2}{c}{Our Eq.~(\ref{eq:m_step_loss})}                      \\ 
Dist.    & Acc.                     & Dist.    & Acc.                         & Dist.    & Acc.                & Dist.    & Acc.                 \\ 
0.0595   & 81.2                     & 0.1521   & 82.2                         & 0.0746   & 88.3                & 0.1680   & 89.2                 \\
\hline
\end{tabular}

}
\label{tab:diversity}
\end{table}
\vspace{10pt}

\subsubsection{Effect of Memory Bank}
\begin{figure}[t!]
    \centering
    \includegraphics[width=0.87\linewidth]{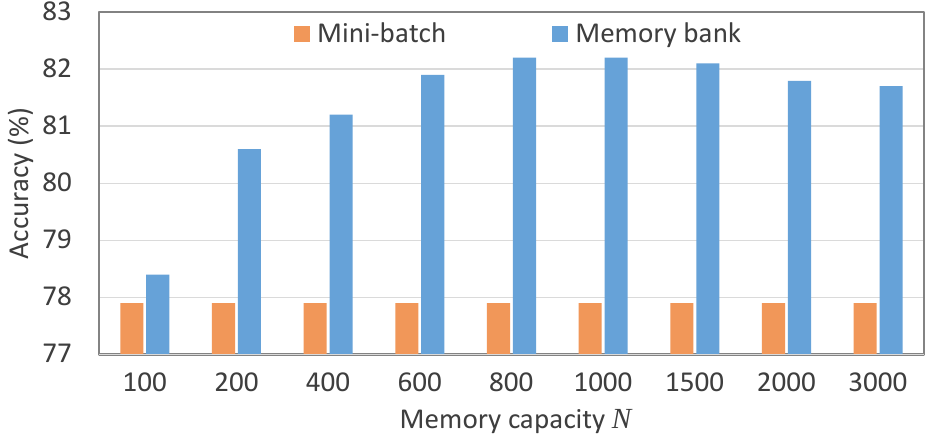}
    \vspace{-12pt}
    \caption{Impact of using memory bank on the classification accuracy, employing a ResNet34-based MGProto model on the CUB dataset. }
    \label{fig:memory}
    \vspace{5pt}
\end{figure}

External memory is used in our MGProto for GMMs to accurately learn the prototype distributions. 
Here, we study the effect of memory capacity $N$ on the classification accuracy. 
The result in Fig.~\ref{fig:memory} shows that the external memory contributes to higher accuracy, compared with the learning from only mini-batch samples.
However, a memory with too large capacity evolves too slowly and the presence of early-trained features can deteriorate performance, 
while a small-capacity memory cannot store enough features for a reliable GMM estimation. 
Hence, we choose the memory capacity $N$ = 800 for CUB 
(for other datasets we select its values based on the class sample ratio relative to CUB).

\subsubsection{$T$ and $\lambda_1$ in Prototype Mining}
\begin{figure}[t!]
    \centering
    \includegraphics[width=0.85\linewidth]{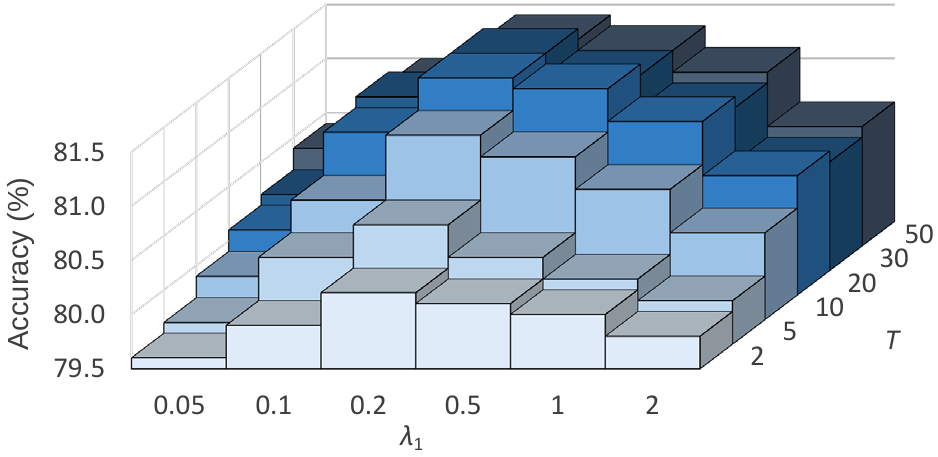}
    \vspace{-12pt}
    \caption{Effects of $T$ and $\lambda_1$ on the classification accuracy of MGProto, evaluated using a ResNet34-based model on CUB. 
    }
    \label{fig:lambda_T}
\end{figure}

We also investigate the the effect of the two important hyperparameters $T$ and $\lambda_1$ utilised in our prototype mining approach, with results shown in Fig.~\ref{fig:lambda_T}. 
It can be seen that when $T$ is greater than 20, the accuracy of MGProto is high and generally stable.
For the mining loss weight $\lambda_1$,  
setting it too low fails to introduce sufficient mining supervision signals, 
whereas increasing it too much also causes decreased classification accuracy. 
Therefore, we have $T=20$ and $\lambda_1 = 0.2$ in all other experiments. 

\section{Discussion and Limitation}

As outlined in Eq.~(\ref{eq:prototype_replacement}) and (\ref{eq:prototype_replacement_my}), the existing prototypical-part networks and our MGProto method require to ground prototypes in the human-understandable space of training images. 
This replacement step aims to find the closest feature patch for each prototype by going through all training images periodically at a regular interval during the training process, as shown in Fig.~\ref{fig:replacement}.
However, this computationally-intensive step can hinder the scalability of these models to datasets with millions of training samples, such as ImageNet~\cite{deng2009imagenet} that has a large number of classes.
Furthermore, the large number of classes may challenge the learning process because prototypes from different classes may become closer to one another, leading to ambiguities in interpretation and an increased risk of misclassification. 
In this regard, developing more efficient replacement strategy (e.g. ball representation of prototypical concepts in ProtoConcept~\cite{ma2024looks}) and more effective learning algorithm for prototypes is highly desirable in future research. 

Our proposed MGProto method leverages class-specific prototypes, which are not shared among classes to ensure clear interpretability for each class.
Nevertheless, this approach may face challenges in scenarios involving a large number of classes (e.g., 1000 classes in ImageNet) due to the following two reasons: 
1) the number of prototypes scales linearly with the number of classes, making it intractable; and
2) as the number of classes grows, the prototypes will become highly similar to each other and the explanation maps they generate for different classes will start to be nearly identical, thereby significantly reducing interpretability. 
One potential solution is to incorporate the idea of class-shared prototypes~\cite{rymarczyk2021protopshare,rymarczyk2022interpretable}, which partially merges prototypes for those classes with similar features. 
By sharing prototypes between classes, we can address scalability (it becomes sublinear in the number of classes) and interpretability (we have many shared prototypes and a few specific prototypes that are more interpretable for each class). 
We argue that this adaptation can be straightforwardly implemented in our method, by making some GMM components shared across related classes.

Our MGProto method evaluates the model interpretability using objective metrics in Table~\ref{tab:interpretability} and Table~\ref{tab:purity}. 
Another possibility to assess the interpretability relies on user studies~\cite{kim2022hive,ma2024looks}, 
which directly gauge how intuitive the explanations appear to human users.
To enhance our evaluation, we plan to  conduct user-based experiments for our proposed method in the future. 
Additionally, considering that our prototypes are learned to model the underlying data distributions, 
it is possible to sample from the distributions and generate new examples for counterfactual explanations. 
We plan to also explore this in future study.

\section{Conclusion}
This work presented MGProto, a generative learning paradigm to produce mixtures of Gaussian-distributed prototypes for interpretable and trustworthy image recognition. 
Different from current point-based prototype learning models, 
MGProto naturally mitigates the performance degradation caused by the projection of point-based learned prototypes, 
and effectively recognises OoD inputs to ensure decision trustworthiness. 
Additionally, a novel and generic strategy is proposed to enhance the prototype learning by mining from less-salient object parts, inspired by the classic legend of Tian Ji’s horse-racing.
MGProto can be substantially compressed via an importance-based pruning strategy, while still keeping high classification accuracy. 
Experimentally, the proposed method obtains SOTA performances on interpretable image recognition, OoD detection, and interpretability quantification. 
A series of ablation studies are also performed to validate the significance of our key claims and effectiveness of the critical model designs.

\section*{Acknowledgements}

This work was supported by funding from the Australian Government under the Medical Research Future Fund - Grant MRFAI000090 for the Transforming Breast Cancer Screening with Artificial Intelligence (BRAIx) Project. G. Carneiro acknowledges the support by the Engineering and Physical Sciences Research Council (EPSRC) through grant EP/Y018036/1.
The authors would like to thank the editor and the anonymous reviewers for their valuable comments and suggestions.











{\small
\bibliographystyle{ieeetr}
\bibliography{refs.bib}
}

 




\vfill



\ifCLASSOPTIONcaptionsoff
  \newpage
\fi

\end{document}


\title{\textit{Supplementary Materials for} \\ Mixture of Gaussian-distributed Prototypes with Generative Modelling for Interpretable and Trustworthy Image Recognition}

\author{Chong Wang,
        Yuanhong Chen,
        Fengbei Liu,
        Davis James McCarthy, 
        Helen Frazer, 
        and Gustavo Carneiro
}


\maketitle

\IEEEdisplaynontitleabstractindextext
\IEEEpeerreviewmaketitle

\section{Additional ablations}
\label{sec:ablation}

vanilla ProtoPNet~\cite{chen2019looks}

\section{Experimental results}

\subsection{Identifying ID and OoD Samples}
\label{sec:IDOoD}

\begin{figure*}[t!]
    \centering
    \includegraphics[width=1.00\linewidth]{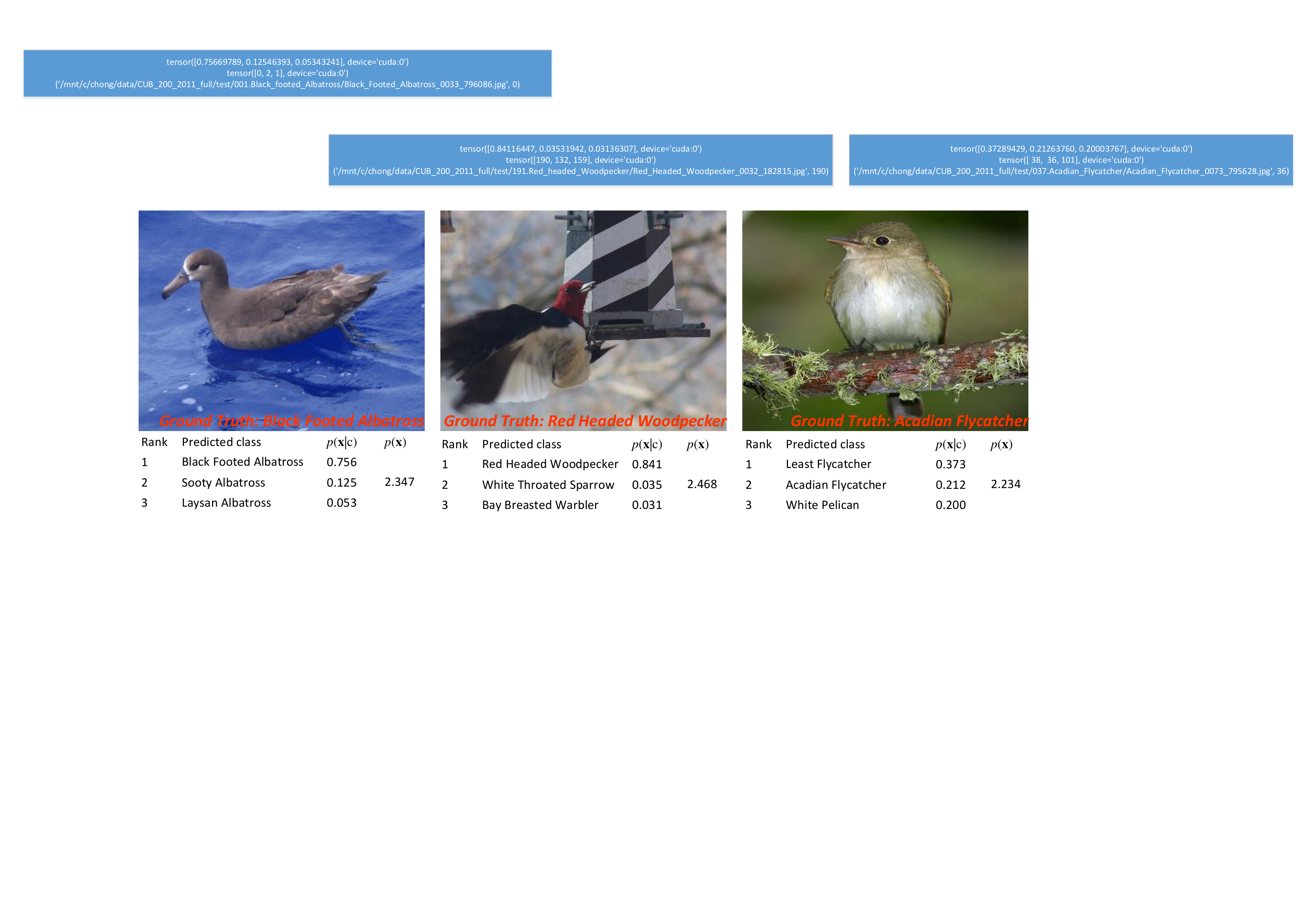}
    \vspace{-20pt}
    \caption{Examples of our MGProto model, trained on CUB-200-2011, for classifying in-distribution (ID) samples from the CUB test set, where we show the predicted top-3 class-conditional probabilities $p(\mathbf{x}|c)$ and overall data probabilities $p(\mathbf{x})$. 
    Our MGProto model outputs high $p(\mathbf{x})$ for these ID samples. The right-hand side image illustrates a misclassified case. 
    }
    \label{fig:ID}
\end{figure*}

\begin{figure*}[t!]
    \centering
    \includegraphics[width=1.00\linewidth]{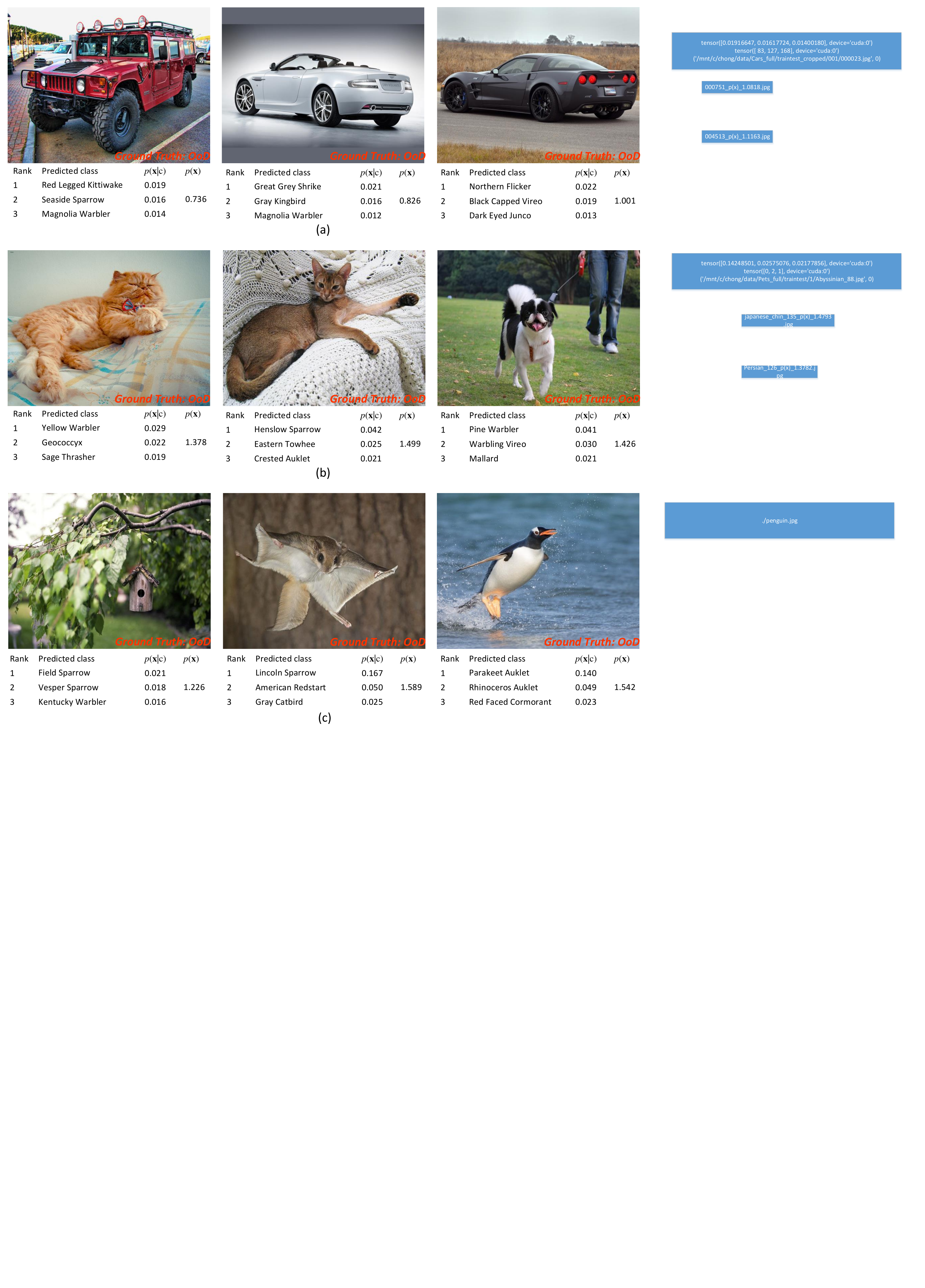}
    \vspace{-15pt}
    \caption{Examples of our MGProto, trained on CUB, for recognising out-of-distribution (OoD) samples from Cars (a), Pets (b), and general images from the Internet (c), where we show the predicted top-3 class-conditional probabilities $p(\mathbf{x}|c)$ and overall data probability $p(\mathbf{x})$. 
    The MGProto model produces low $p(\mathbf{x})$ for these OoD samples. 
    Note in (c) that the left-hand side image contains only tree branches on which birds often perch, the middle and right-hand side images are near-distribution samples containing a Flying Squirrel and a Penguin, respectively. 
    }
    \label{fig:OoD}
\end{figure*}

In Fig.~\ref{fig:ID} and Fig.~\ref{fig:OoD}, we illustrate typical classification predictions from our MGProto, 
for classifying In-Distribution(ID) samples and identifying Out-of-Distribution (OoD) inputs, respectively.
Our model is trained on CUB-200-2011. 

Fig.~\ref{fig:ID} showcase that our method produces large data probability $p(\mathbf{x})$ for these CUB images, 
which are higher than the OoD detection threshold (denoted by star in the Fig. 7 of main paper). 
Hence, these images are treated as ID samples and the interpretable classification decisions made on them can be deemed reliable and worthy of acceptance.

By contrast, the OoD inputs in Fig.~\ref{fig:OoD} obtain low $p(\mathbf{x})$ (smaller than the OoD detection threshold).
Therefore, the model opts to abstain from classifying these input images because they do not fit well the CUB-prototype distributions of all 200 classes.
In Fig.~\ref{fig:OoD}(c), the left-hand side image contains only tree branches on which birds often perch, and the model identifies it as OoD, 
showing that our method can successfully distinguish images without containing any object, even when accompanied by highly relevant contextual information. 
The middle and right-hand side images in Fig.~\ref{fig:OoD}(c) are near-distribution OoD samples. 
For instance, the penguin exhibits certain similarities, such as its white belly, to the Parakeet Auklet (one of the 200 classes in CUB), resulting in the model assigning a relatively large $p(\mathbf{x}|c)$ for the class Parakeet Auklet, as well as a relatively large $p(\mathbf{x})$. 
However, despite this near-distribution nature, the model is still capable of recognising these samples.

{\small
\bibliographystyle{ieeetr}
\bibliography{refs}
}

 




\vfill



\ifCLASSOPTIONcaptionsoff
  \newpage
\fi